\documentclass{article} 
\usepackage[preprint]{colm2026_conference}

\usepackage{microtype}
\usepackage{hyperref}
\usepackage{url}
\usepackage{booktabs}

\usepackage{lineno}

\definecolor{darkblue}{rgb}{0, 0, 0.5}
\hypersetup{colorlinks=true, citecolor=darkblue, linkcolor=darkblue, urlcolor=darkblue}
\newcommand{\authgap}{\nobreak\hspace{0.5em}}

\title{ProofVerifier: A Scalable, Diversity-Driven Framework for Natural-Language Proof Verification}

\author{
Haotong Yang\thanks{
    Equal contribution.
    $^\dagger$Corresponding authors.
    $^\ddagger$Work at Tencent.
    Correspondence to:
    Zhouchen Lin \texttt{<zlin@pku.edu.cn>},
    Muhan Zhang \texttt{<muhan@pku.edu.cn>}.
}\authgap$^{\ddagger 1}$,
Zitong Wang\footnotemark[1]\authgap$^{1}$,
Shijia Kang$^{1}$,
Siqi Yang$^{1}$,
Wenkai Yu$^{1}$,
Xu Niu$^{1}$, \\[0.1em] \bfseries
Yike Sun$^{1}$,
Yi Hu$^{1}$,
Zhouchen Lin$^{\dagger 1}$,
Muhan Zhang$^{\dagger 1}$
\\[0.3em]
$^{1}$Peking University, Beijing, China
}

\usepackage{hyperref}
\usepackage{url}
\usepackage{booktabs}
\usepackage{multirow}
\usepackage{graphicx}
\usepackage{amssymb}
\usepackage[dvipsnames]{xcolor}
\usepackage{caption} 
\usepackage{ulem}
\usepackage{enumitem}
\usepackage{tabularx}
\usepackage{array}
\usepackage{makecell}   
\usepackage{arydshln}
\usepackage{wrapfig}
\usepackage{xcolor}
\usepackage{listings}
\usepackage[many]{tcolorbox}
\usepackage{hyperref}
\usepackage{nameref}
\usepackage{CJKutf8}
\newcolumntype{L}{>{\raggedright\arraybackslash}m{0.25\textwidth}}
\newcolumntype{C}{>{\centering\arraybackslash}m{0.18\textwidth}}
\newcolumntype{R}{>{\raggedright\arraybackslash}m{0.35\textwidth}}

\usepackage{amsmath,amsfonts,bm}

\def\eqref#1{equation~\ref{#1}}

\def\1{\bm{1}}

\DeclareMathAlphabet{\mathsfit}{\encodingdefault}{\sfdefault}{m}{sl}
\SetMathAlphabet{\mathsfit}{bold}{\encodingdefault}{\sfdefault}{bx}{n}

\definecolor{c1}{HTML}{DDEAF6}
\definecolor{c2}{HTML}{5A9AD4}
\definecolor{mygray}{gray}{0.6}

\newtcolorbox[use counter=task,
              number within=section,
              number freestyle={\thesection-\noexpand\arabic{\tcbcounter}},]{answer}[1][]{
    colback=c1!40,
    colframe=c2,
    fonttitle=\bfseries,
    coltitle=white,
    breakable,
    #1 
}
\newtcolorbox[use counter=task,
              number within=section,
              number freestyle={\thesection-\noexpand\arabic{\tcbcounter}},]{promptbox}[1][]{
    colback=mygray!40,
    colframe=mygray,
    fonttitle=\bfseries,
    coltitle=white,
    breakable,
    #1 
}

\begin{document}

\ifcolmsubmission
\linenumbers
\fi

\maketitle

\begin{abstract}
While large language models (LLMs) have achieved strong performance on mathematical problems with verifiable answers, many advanced problems are proof-based and require evaluating full proofs. However, training such verifiers requires diverse and trustworthy question-proof-check (QPC) examples at scale, which are scarce. To address this challenge, we develop a human-audited, LLM-assisted data pipeline that produces large-scale QPC triplets with limited human effort. By systematically varying problem sources, generation strategies, and generator models, the pipeline creates diverse problem-proof pairs spanning multiple difficulty levels, linguistic styles, and error types. We combine multi-LLM agreement with hierarchical human auditing to obtain accurate proof-correctness labels. Using these data, we train generative proof verifiers and introduce an auxiliary fluency filter together with balanced token weighting to stabilize binary-reward long-form verification RL. Experiments show that our verifier improves proof-judgment accuracy across different proof styles and provides useful guidance for test-time selection. Overall, our results provide a practical data and training framework for natural-language proof verification.
\end{abstract}

\section{Introduction}
\label{sec:intro}
Large language models (LLMs) are making rapid progress on mathematical reasoning, especially on tasks whose final answers can be checked cheaply by exact match, symbolic tools, or lightweight judging~\citep{Guo2025-qu,deepseekai2025deepseekr1incentivizingreasoningcapability,huang2025gemini25procapable}. Many advanced mathematical tasks, however, are proof-based: a final claim may be correct even when the proof used to derive it is invalid. In this setting, verification must evaluate the argument itself rather than only the endpoint. Practical Natural-Language (NL) proof verification is therefore important for evaluation, test-time selection, and reinforcement learning on proof generation.

In view of the intrinsic difficulty and flexibility of proof verification tasks, it is natural to consider finetuning LLMs as proof verifiers~\citep{dekoninck2025openproofcorpuslargescale}, \citep{shao2025deepseekmathv2selfverifiablemathematicalreasoning}.
Finetuning such proof verifiers needs question-proof-check (QPC) triples, where each example contains a problem, a candidate proof, and a proof-level correctness judgment, and where both valid and invalid proofs are well represented. Official or reference solutions provide high-quality positive proofs, but they do not by themselves supply the incorrect proofs needed to train a practical verifier. Obtaining such negatives with reliable labels through full expert annotation is expensive, which limits scalability. This challenge is compounded by the \textit{heterogeneity} of proof verification itself. Correct proofs vary in structure, detail level, and presentation style, while incorrect proofs span logical gaps, theorem misuse, hallucinated claims, and incomplete arguments. Therefore, we identify the absence of a data pipeline that is both \textit{scalable} and \textit{designed for broad proof coverage} as a key bottleneck.

\begin{figure*}[t]
    \centering
    \includegraphics[width=0.95\linewidth]{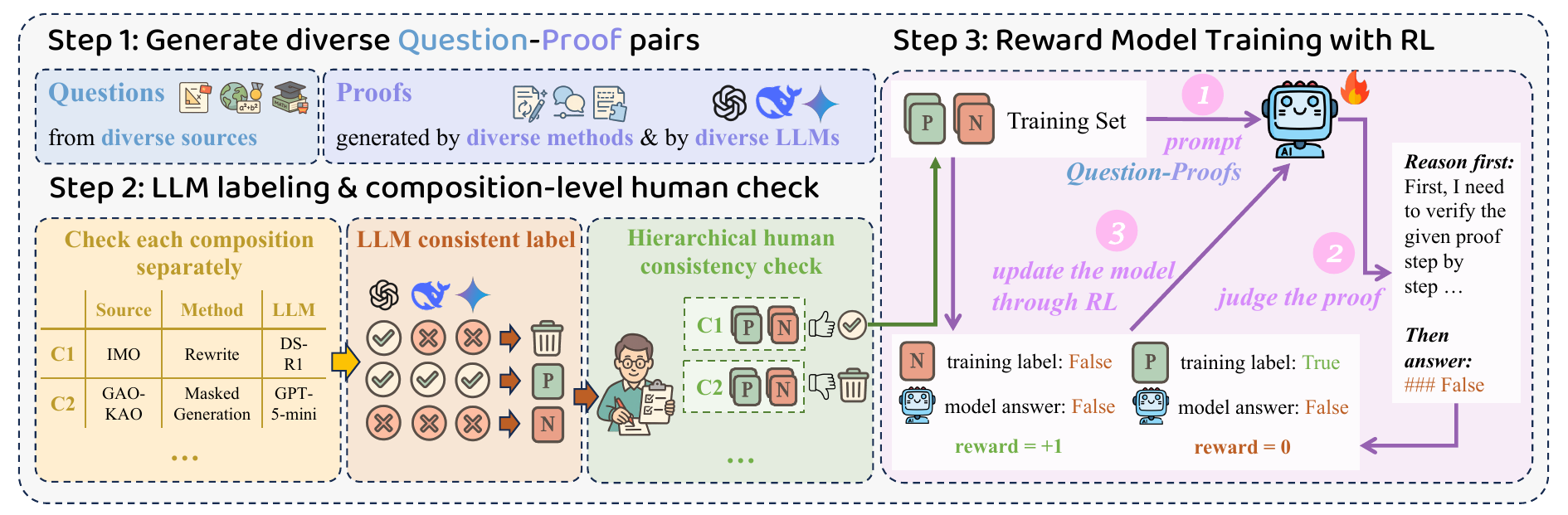}
    \vspace{-5pt}
    \caption{Pipeline of data construction and verifier training in our framework.}
    \vspace{-10pt}
    \label{fig:big_pipeline}
\end{figure*}

We address this bottleneck with our framework, a human-audited, LLM-assisted data pipeline for natural-language proof verification that produces the ProofVerifier models (Figure~\ref{fig:big_pipeline}). At a high level, the framework first broadens proof data coverage along three axes---problem source, proof-generation strategy, and generator model---so that the verifier is exposed to substantially more varied positive and negative proofs.
To obtain reliable human-audited silver labels for these proof candidates at scale, we then combine multi-LLM agreement with hierarchical human auditing, using selective human auditing rather than full item-level manual annotation. We curate more than 20,000 training QPC triples, with the accepted slices achieving an average of 94.3\% agreement with human judgments during the human auditing process. Our pipeline also reduces human annotation effort by more than 12x relative to full manual checking.

Given such data, training a generative proof verifier yields a setting that is not well captured by current standard practice of RL with verifiable reward (RLVR). We call this setting \textit{binary-reward} long-form verification RL: given a question-proof pair, the verifier generates a long free-form verification trace together with a final binary verdict, while the training reward depends only on whether that final verdict matches the proof-correctness label. The difficulty is that supervision is binary but reasoning is long-form. We find two main sources of instability: binary rewards can reinforce noisy early behaviors because even a random model receives positive feedback 50\% of the time, and standard GRPO~\citep{deepseekai2025deepseekr1incentivizingreasoningcapability} token weighting introduces length bias that is especially severe when the reward can be cheaply exploited.

To tackle such instabilities present in the RL setting, we introduce balanced token weighting to control length bias and an auxiliary LLM-based fluency judge to suppress anomalous reasoning traces. With our stable algorithm, we train the ProofVerifier models with more than 300 RL steps and covering over 18k training samples across 8B, 14B, and 32B models.

Our experiments show that the resulting verifier is both accurate and practically useful. The ProofVerifier models improve proof-judgment accuracy on our curated test set, stay competitive on OPC and ProofBench, remain robust under style shifts, and provide useful best-of-$k$ guidance for test-time selection. 

Our contributions are:
\begin{itemize}[leftmargin=10pt,topsep=0pt,itemsep=0pt,parsep=0pt]
    \item A scalable, diversity-driven QPC data-construction pipeline that expands proof coverage across sources, prompting methods, and generator models.
    \item A multi-LLM agreement plus hierarchical human-audit procedure that controls label quality while reducing annotation effort by more than 12x.
    \item A stable training recipe for binary-reward long-form verification RL, centered on balanced token weighting and an auxiliary LLM-based fluency judge.
\end{itemize}

\section{Related Work}
\label{sec:related_work}

\paragraph{Reinforcement learning with verifiable reward} Recent progress in mathematical reasoning has been driven in large part by RLVR~\citep{deepseekai2025deepseekr1incentivizingreasoningcapability,lambert2025tulu3pushingfrontiers,huang2025gemini25procapable}, which optimizes language models against automatically verifiable reward signals provided by external checkers rather than human preferences.  These methods work especially well when correctness can be checked cheaply from the final answer. Proof-based tasks are harder because the reward signal must evaluate the full argument rather than only the endpoint. Some prior works attempt to reduce proofs to more easily verifiable subproblems~\citep{balunović2025mathconstructchallengingllmreasoning,sheng2025solvinginequalityproofslarge}, and others rely on professional mathematicians to create problems with such endpoints, nevertheless exhibiting issues of domain generalization and scalability.  

\paragraph{Proof verification models} Recent work has studied proof verification models~\citep{dekoninck2025openproofcorpuslargescale,zheng2023judgingllmasajudgemtbenchchatbot,li2023generativejudgeevaluatingalignment,mahan2024generativerewardmodels,zhang2025generativeverifiersrewardmodeling, ma2025reliablefinegrainedevaluationnatural}, exploring whether the LLM can verify proof correctness. 
OPC~\citep{dekoninck2025openproofcorpuslargescale} provides a fully human-labeled corpus for proof generation and verification together with a strong proof judge. ProofBench~\citep{ma2025reliablefinegrainedevaluationnatural} provides insight into evaluating and eliciting the ability of LLMs to generate fine grained verification results. 
Both datasets rely on substantial human annotation, which limits the scalability of training-data creation. To solve the problem, we build a large human-audited silver corpus for verifier training and pair it with a practical training recipe for binary-reward long-form verification RL. 


\section{Data Pipeline}

\begin{figure*}[t]
    \centering
    \includegraphics[width=0.85\linewidth]{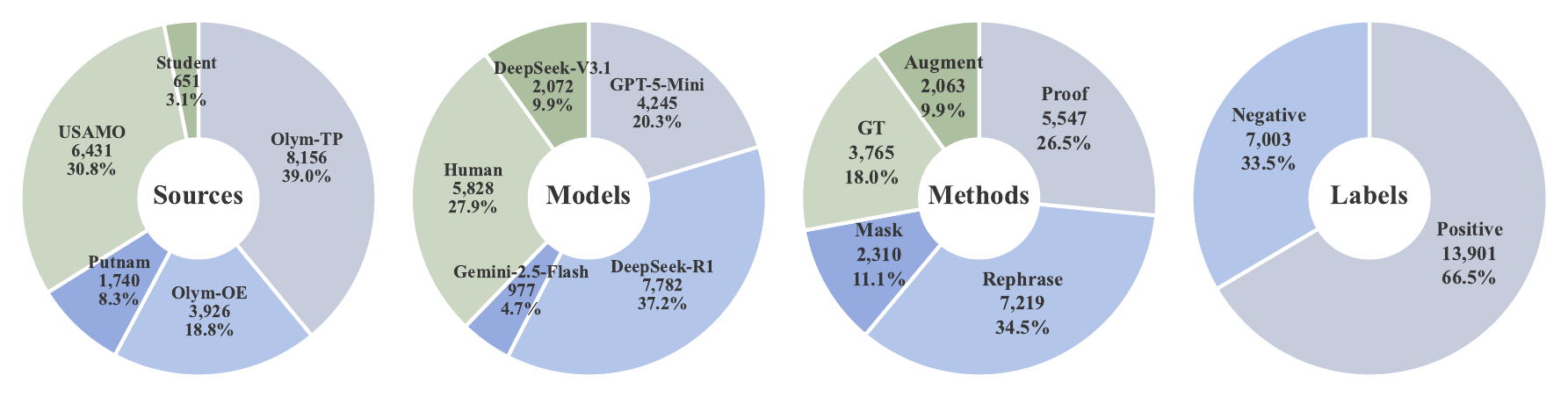}
    \caption{Training data distribution generated by the data pipeline.}
    \label{fig:data_distribution}
    \vspace{-5pt}
\end{figure*}

\subsection{Automated Diverse Data Collection}
\label{ssec:qa_collection}
Proof verification is highly heterogeneous: correct proofs vary widely in structure, detail, and style, while incorrect proofs exhibit diverse logical and mathematical errors. A robust verifier therefore requires training data that covers varied problem settings, proof styles, reasoning patterns, and failure modes.

Concretely, we expand the corpus along three axes: problem source, proof-generation method, and generator model. We denote the data generated under one source-method-model combination as a \textit{slice}.

\paragraph{Question-proof source diversity}
Competition-level proof problems form the core of our corpus because they require substantial reasoning and are accompanied by curated reference solutions. We include problems from OlympiadBench~\citep{he2024olympiadbench} (TP and OE split), USAMO, and the Putnam Competition. These sources vary not only in difficulty, but also in problem format, expected proof style, and the structure of the corresponding reference solutions. Overall, the resulting collection spans the period from 1985 to 2025 and covers a broad range of difficulty, from standard high-school olympiad problems to challenging international contest problems. The reference solutions are likewise highly heterogeneous, ranging from terse writeups of a few hundred tokens to detailed expositions exceeding 2,000 tokens.
Details of the adopted source construction, including the open-ended setting, are provided in Appendix~\ref{app:question_problem_source}. 

\paragraph{Proof-generation method diversity}
To broaden both positive styles and negative error modes, we use LLMs to generate additional proofs from the seed problems. Different prompting strategies induce markedly different proof behaviors, as indicated in Table~\ref{tab:prompt_summary}
This axis is especially important because realistic negative proofs are otherwise scarce in reference-solution-based resources. More examples of the generated error types are shown in Appendix~\ref{app:negative_sample_analysis}. We also add a small amount of auxiliary data, including short-answer computational problems and deliberately degenerate proofs such as refusals or vacuous outputs. These examples improve coverage of easy-to-detect negatives and simple boundary cases.

\paragraph{Generator-model diversity}
The choice of generator model introduces further variation in proof length, tone, theorem usage, and characteristic error patterns. As a practical trade-off between diversity and generation cost, we use DeepSeek-R1, DeepSeek-V3.1, GPT-5-mini, and Gemini-2.5-Flash to produce solutions with distinct stylistic tendencies. For example, DeepSeek-R1 often produces more conversational, human-like writeups, compared to the more explicitly structured solutions of other models.

Together, these three axes extend coverage well beyond reference solutions alone, especially for incorrect proofs and stylistic variation among valid ones. In Section~\ref{ssec:ablation}, we examine the empirical contribution of these design choices through diversity ablations.

\subsection{LLM-aided Consistent Labeling with Slice-level Human Check}
\label{ssec:llm_annotation}
Once QP pairs are collected, we need reliable binary labels indicating whether each proof is correct. Full expert annotation for every sample is prohibitively expensive, so we use a two-stage protocol designed to produce reliable human-audited silver labels while sharply reducing human effort.

First, each proof is judged five times by three frontier LLMs: DeepSeek-R1 is queried three times, GPT-5-mini once, and Gemini-2.5-Flash once. We keep a sample only if all five judgments agree. This unanimity filter is intentionally conservative: it sacrifices raw volume in order to retain only those examples whose provisional labels are already stable across repeated checks and across models before any human audit is applied.

Second, we audit the retained data at the \textit{slice level}. Each slice is defined by a combination of problem source, prompting method, and generator model, so that it corresponds to one concrete generation regime. We use this unit because samples produced under the same source-method-model setting tend to share similar labeling behavior and recurring failure modes, making regime-level reliability the central question.

For each slice, we first generate a pilot pool by sampling 5\% of the source questions and collecting the associated QP pairs. We then conduct three cumulative rounds of human checking by subsampling from this pilot pool, bringing the total fraction of QP pairs audited in that slice to approximately 0.5\%, 1\%, and 2.5\% over the three rounds. The required agreement between LLM consensus and human judgments is 75\% in the first round, 80\% in the second, and 90\% in the third. In addition, we require at least 30 checked proofs for each audited slice to avoid making slice-level decisions from trivially small audits. At the minimum acceptance threshold, this protocol implies with 90\% confidence that the average consistency rate within a accepted slice is above 79.1\% using the Clopper-Pearson lower bound, which is acceptable in a large-scale RL scenario.

During the auditing regime, if a slice fails any round, we discard the entire slice. Otherwise, we carry out the full generation process for that slice and keep its unanimous LLM labels as human-audited silver labels for training. This design directs human effort toward validating the reliability of a generation regime rather than re-annotating every item independently, which is exactly where large-scale savings arise.

In total, we audit 30 slices and accept 18 of them. The accepted slices average 94.3\% consistency with human judgments. Our final training set contains 27 combinations and 20,904 QPC triples. The full distribution is shown in Figure~\ref{fig:data_distribution}, with a detailed breakdown in Appendix~\ref{app:data_distribution}. The human audition details are available in Appendix~\ref{app:detail_of_human_check}.

This protocol also yields a substantial efficiency gain. As detailed in Appendix~\ref{app:human-check-detail}, fully manual annotation for the 20k-scale training set would require at least 1,200 hours. By contrast, our slice-level audit process required less than 100 human hours in actual implementation, corresponding to more than a 12x reduction in annotation effort.

For evaluation, we build a human-annotated internal test set by holding out audited questions at the question level and using the labeled proofs associated with those questions, while keeping both LLM-consistent and LLM-inconsistent proofs. It contains 238 question-proof pairs from 51 questions, with 141 positive and 97 negative examples, with all QP pairs double-labeled by human annotators. We also utilze test sets in OPC~\citep{dekoninck2025openproofcorpuslargescale}, ProofBench~\citep{ma2025reliablefinegrainedevaluationnatural} as indicators for transfer to external benchmarks. We provide a detailed description of our test set in Appendix~\ref{app:test_set_detailes}.

\begin{table*}[tb]
\small
\centering
\resizebox{1.0\linewidth}{!}{
\begin{tabular}{@{}ccll@{}}
\toprule
\multicolumn{1}{c}{\textbf{Name}} &
\multicolumn{1}{c}{\textbf{Input}} &
\multicolumn{1}{c}{\textbf{Method}} &
\multicolumn{1}{c}{\textbf{Proof Behavior}} \\
\midrule

Rephrase &
QP &
\makecell[l]{Rephrase the proof \\in the model's own style} &
\makecell[l]{A model-generated rewrite of a reference proof; \\usually correct but may introduce local step-level flaws} \\ \addlinespace[3pt] \hdashline
\addlinespace[3pt]

Proof &
Q &
\makecell[l]{Construct a proof by itself} &
\makecell[l]{A model-generated proof attempt with standard heuristic ideas; \\common flaws include case-analysis errors or hallucinated lemmas} \\ \addlinespace[3pt] \hdashline
\addlinespace[3pt]

\makecell[c]{Mask\\Completion} &
\makecell[c]{Q \&\\Masked P} &
\makecell[l]{Split the proof, mask key steps,\\ and let the LLM fill them in} &
\makecell[l]{A reference-style proof with subtle local omissions \\or unjustified transitions; errors are harder to notice \\and closer to human step-level mistakes} \\ \addlinespace[3pt] \hdashline
\addlinespace[3pt]

Augment &
QP &
\makecell[l]{Apply incremental wording or language\\changes while preserving content} &
\makecell[l]{An incrementally perturbed reference-style proof \\retaining the original reasoning and any existing errors} \\

\bottomrule
\end{tabular}
}
\caption{Different input and prompt methods induce distinct proof behaviors.}
\label{tab:prompt_summary}
\end{table*}

\section{Stable Training for ProofVerifier}

\label{sec:stable_training}

We use the curated QPC data to train a generative verifier. Our base models are DeepSeek-R1-0528-distill-Qwen3 at 8B, 14B, and 32B scale. We adopt a generative reward-model formulation to exploit their ability to reason about logical structure while usable for downstream tasks.
We implement training in \textbf{verl}~\citep{sheng2024hybridflow} and use GSPO~\citep{zheng2025groupsequencepolicyoptimization} as the base RL algorithm, with hyperparameters detailed in Appendix~\ref{app:rl_hyper_param}. 

\subsection{Balanced Token Weighting}
\label{ssec: balanced token weighting}
One major source of instability is how token-level credit is normalized across responses of different lengths. In proof verification, output length varies naturally: some cases can be resolved concisely, whereas others require much longer analysis to localize the decisive flaw or justify why a proof is valid. Under binary reward, this variation becomes especially consequential, because many responses with substantially different reasoning quality can still share the same sequence-level signal. The resulting training dynamics therefore depend not only on which responses are rewarded, but also on which kinds of rewarded responses are amplified most strongly.

Let \(o_1,\dots,o_G\) denote the group of rollout outputs for one prompt, let \(A_i\) be the sequence-level advantage of rollout \(o_i\), and let \(|o_i|\) be the generated-token length of output \(o_i\). We write the policy-gradient term in the form
\[
\mathcal{L}(\theta)
=
-\sum_{i=1}^{G}\sum_{t=1}^{|o_i|}
w_i\,\ell_{i,t}^{\mathrm{clip}}(\theta),
\]
where \(w_i\) is the response-level token weight introduced by normalization schemes, and
$
\ell_{i,t}^{\mathrm{clip}}(\theta)
=
\min(
r_{i,t}(\theta)A_i,\;
\mathrm{clip}(r_{i,t}(\theta),1-\epsilon,1+\epsilon)A_i
)
$
is the standard clipped surrogate term with
$
r_{i,t}(\theta)
=
\frac{\pi_\theta(o_{i,t}\mid q,p,o_{i,<t})}
{\pi_{\theta_{\mathrm{old}}}(o_{i,t}\mid q,p,o_{i,<t})}.
$
The distinction studied in this section comes from the choice of \(w_i\). Two standard choices allocate update mass very differently. Under a GRPO-like~\citep{deepseekai2025deepseekr1incentivizingreasoningcapability} normalization,
$
w_i^{\mathrm{GRPO}}=\frac{1}{G|o_i|},
$
so each response receives roughly the same total update mass regardless of length. Under a DAPO-like~\citep{yu2025dapoopensourcellmreinforcement} normalization,
$
w_i^{\mathrm{DAPO}}=\frac{1}{\sum_{j=1}^G |o_j|},
$
so tokens from different responses receive comparable per-token weight and longer responses accumulate larger total update mass. These two schemes therefore induce different optimization tendencies with respect to response length, and in our setting they fail in opposite directions.

For GRPO-like weighting, the key issue is the response-length bias highlighted by Dr.~GRPO~\citep{liu2025understandingr1zeroliketrainingcritical}: positively advantaged short responses receive relatively stronger updates, while negatively advantaged long responses are corrected less aggressively. In our task, this bias interacts with early-stage model behavior in a particularly harmful way. Importantly, at the beginning of training, verifiers are often permissive, prefer accepting a proof rather than reliably identifing the decisive error. These early acceptance responses are much shorter than diagnostic rejection responses that require explicit flaw localization. As a result, once such short superficial acceptance traces receive positive advantage, GRPO-like weighting amplifies them disproportionately. The policy is then encouraged to reinforce an early shortcut instead of learning reliable proof diagnosis.

DAPO-like weighting alleviates this short-positive/long-negative asymmetry, but it introduces a different failure mode in our setting. Because longer responses accumulate larger total update mass, the policy is encouraged to spend more tokens before reaching the final verdict. The result is a strong incentive toward increasingly long responses, often accompanied by repetitive or low-information text. Thus, while GRPO-like weighting tends to over-amplify short rewarded acceptance traces early in training, DAPO-like weighting tends to over-reward long and often redundant responses.


These complementary failure modes motivate our balanced token weighting:
\[
w_i^{\mathrm{bal}}
=
\eta\frac{1}{G|o_i|}
+
(1-\eta)\frac{1}{\sum_{i=1}^G |o_i|}.
\]
This interpolation trades off between per-sample and global normalization. The first term preserves enough per-sample normalization to prevent long responses from dominating solely because they contain more tokens. The second term preserves enough global normalization to avoid over-amplifying very short rewarded traces. 

In our experiments, \(\eta=0.6\) is an empirical choice that works consistently across the tested 8B, 14B, and 32B models, stabilizes output length, and improves training reliability.  We therefore view balanced token weighting as a targeted fix for a characteristic failure mode of binary-reward long-form verification RL. Section~\ref{ssec:ablation} reports both reward and output-length behavior when ablating the usage of different weighting schemes.

\subsection{LLM-based Fluency Judge}
The second instability source is pathological trace quality. In our setting, the RL signal depends only on whether the final verdict is correct, not on the local quality of the generated verification trace. As a result, a malformed trace can still receive positive reward if it happens to end with the correct binary judgment. This creates a failure mode in which training may reinforce poor reasoning behavior rather than only correct verification decisions.

In practice, model collapse is often preceded by an early collapse signature: the verifier begins to emit obviously odd local behavior---repeated phrases, self-interruptions, malformed meta-comments, looping transitions, or abrupt verdicts with little justification---while still occasionally landing on the correct final verdict. Some of these anomalies can arise stochastically during rollout and would not matter much if they were never rewarded. The problem is that once such traces receive positive reward, they can become self-reinforcing: what begins as occasional noise is turned into learned behavior, the behavior appears more frequently in later rollouts, and the policy can rapidly drift toward degenerate generation. In the collapse cases we inspected, once these precursor traces became visible, severe degradation often followed within roughly 5--10 optimization steps if they were not filtered.

To address this, we add an auxiliary LLM-based fluency judge that checks the verifier's generated reasoning for surface-level anomalies. The judge is not asked to re-solve the proof problem or reassess the mathematical correctness of the verdict. Instead, it focuses on behavioral pathologies in the generated trace itself: repetition, local loops, malformed meta-commentary, irrelevant insertions, and abruptly asserted conclusions unsupported by the preceding reasoning. If such anomalies are detected, the sample is treated as invalid for reward purposes regardless of the final verdict. This makes the module complementary to the main verifier: the verifier is responsible for mathematical judgment, while the fluency judge suppresses traces that are behaviorally unfit to reinforce.

We instantiate this auxiliary module with DeepSeek-V3.1-chat, with deliberate reasoning disabled, because the task is behavioral rather than mathematical. In one sampled training interval, the fluency judge flagged 0.6\%--2.0\% of outputs as anomalous; manual inspection gave approximately 96\% true positive and 96\% true negative rates. These numbers suggest that the judge acts as a sparse but meaningful filter: it intervenes on only a small fraction of rollouts, yet those interventions are accurate enough to remove exactly the kinds of traces that can act as collapse precursors. Utilizing the one-step-delay training scheme in \textbf{verl}, we only introduce moderate overhead, with the details in Appendix~\ref{app:overhead}.

With these modifications, training remains stable for more than 300 RL steps, covering over 18k training samples and 112M generated tokens.

\begin{wrapfigure}[7]{R}{0.4\linewidth}
\centering
\vspace{-25pt}
\small
    \includegraphics[width=\linewidth]{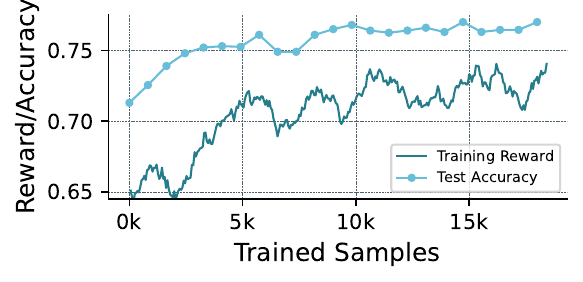}
    \vspace{-10pt}
    \caption{Training curve of ProofVerifier.}
    \vspace{-10pt}
    \label{fig:training_curve}
\end{wrapfigure}

Figure~\ref{fig:training_curve} summarizes the overall training behavior induced by the method described above. The average reward rises together with held-out accuracy over the entire training run, indicating the effectiveness of our training recipe.

\section{Experiments}
\label{sec:experiment}
We evaluate the ProofVerifier models from four angles: verifier accuracy on our held-out human-annotated test set, robustness on external and distribution-shifted evaluations, ablations on the data pipeline and training recipe, and best-of-$k$ reranking for test-time selection. Unless otherwise specified, all verifier accuracies are reported as Avg@8, taking an average over 8 independent rollout samples.

\begin{table}[t]
\centering
\small

\begin{minipage}{0.45\textwidth}
\begin{tabular}{ll}
\toprule
\textbf{Model} & \textbf{Accuracy} \\ \midrule
DeepSeek V3.1         & 69.6 \\
Gemini-2.5-Flash      & 74.5 \\
GPT-5-mini-2025-08-07 & 74.6 \\ \midrule
Distill-Qwen3-8B      & 71.6 \\
Distill-Qwen3-14B     & 71.9 \\
Distill-Qwen3-32B     & 72.5 \\
ProofVerifier-8B      & 76.8 {\footnotesize (+5.2)} \\
ProofVerifier-14B     & 79.0 {\footnotesize (+7.1)} \\
ProofVerifier-32B     & \textbf{82.4} {\footnotesize (+9.9)} \\ \bottomrule
\end{tabular}
\caption{Comparison on the held-out human-annotated test set. Numbers in parentheses denote gains over base models.}
\label{tab:main_results_iid}
\end{minipage}
\hfill
\begin{minipage}{0.52\textwidth}
\begin{tabular}{lcc}
\toprule
\textbf{Model} & \textbf{OPC} & \textbf{ProofBench} \\ \midrule
o3                & 83.1 & --   \\
Gemini-2.5-Flash  & 82.7 & --   \\
DeepSeek-R1       & 80.9 & --   \\ \midrule
Distill-Qwen3-8B  & 70.8 & 59.3 \\
Distill-Qwen3-14B & 69.4 & 59.8 \\
Distill-Qwen3-32B & 70.2 & 56.9 \\
ProofVerifier-8B  & 75.7 {\footnotesize (+4.9)} & 67.5 {\footnotesize (+8.2)} \\
ProofVerifier-14B & 77.4 {\footnotesize (+8.0)} & 68.5 {\footnotesize (+8.7)} \\
ProofVerifier-32B & 81.3 {\footnotesize (+11.1)} & 69.4 {\footnotesize (+12.5)} \\ \bottomrule
\end{tabular}
\caption{Transfer results on OPC and ProofBench.}
\label{tab:transfer_results}
\end{minipage}
\vspace{-18pt}
\end{table}

\subsection{Verifier Accuracy on the Curated Test Set}
We utilize the test set constructed as indicated in  Section~\ref{ssec:llm_annotation}. Table~\ref{tab:main_results_iid} shows consistent gains over the base models at all scales. ProofVerifier-8B is already competitive with the closed-source judges, and ProofVerifier-32B reaches 82.4 Avg@8. This indicates that the training recipe transfers well across model scale.

To obtain a coarse diagnostic view of what kinds of incorrect proofs are easier or harder for the verifier to reject, we manually analyze the 97 negative question-proof pairs in this test set. We group them into three common failure modes, as detailed in Appendix~\ref{app:test_set_detailes}. These categories are intended as a compact empirical diagnostic, chosen because they recur frequently in manual inspection and provide an interpretable breakdown of where proofs fail. Table~\ref{tab:error_type_breakdown} shows that the ProofVerifier-8B model improves across all three error types, serving as semi-quantitative evidence that the trained verifier is capable of identifying diverse types of error. Appendix~\ref{app:error_case_study} retains longer qualitative case studies of these behaviors.

\begin{table}[t]
  \centering
  \small
  \begin{tabular}{@{}lrrr@{}}
\toprule
\textbf{Error Type}                         & \textbf{Percentage} & \textbf{Base Model Acc} & \textbf{ProofVerifier Acc} \\ \midrule
Misusing unjustified assumptions   & 15.5 \%                         & 53.3           & 61.7 (+8.4)       \\
Incomplete logic flow              & 56.7 \%                         & 42.7           & 56.8 (+14.1)      \\
Making up non-existent conclusions & 27.8 \%                         & 41.3           & 55.6 (+14.3)      \\ \bottomrule
\end{tabular}
  \caption{Accuracy breakdown by error type on a manually categorized negative subset.}
  \label{tab:error_type_breakdown}
\end{table}

\subsection{Transfer and Robustness}
We next evaluate whether the verifier transfers beyond our curated benchmark. We consider two externally constructed proof-verification benchmarks, OPC~\citep{dekoninck2025openproofcorpuslargescale} and ProofBench~\citep{ma2025reliablefinegrainedevaluationnatural}, together with three style-shifted variants of our held-out test set. The full test set details are available in Appendix~\ref{app:test_set_detailes}

As is shown in Table~\ref{tab:transfer_results}, the ProofVerifier models demonstrates universal performance gain over the base models across the two test sets. The same pattern also strengthens with the size of the models. Together, these results show that the gains from our training pipeline carry over to independently constructed proof-verification evaluations. 

\begin{table}[t]
\centering
\small
\begin{minipage}{0.45\textwidth}
\begin{tabular}{lccc}
\toprule
\textbf{Model} & \textbf{Curated} & \textbf{OPC} & \textbf{Avg} \\ \midrule
Distill-Qwen3-8B & 71.6 & 70.8 & 71.2 \\
OPC-R1-8B*        & 64.6 & 82.0 & 73.3 \\
ProofVerifier-8B & \textbf{76.8} & 75.7 & \textbf{76.2} \\ \bottomrule
\end{tabular}
\vspace{-5pt}
\caption{Comparison across our curated test set and OPC.*: The results are obtained by us testing on the official model weights.}
\label{tab:head_to_head_opc}
\end{minipage}
\hfill
\begin{minipage}{0.5 \textwidth}
\begin{tabular}{lccc}
\toprule
\textbf{Proof Style} & \textbf{Base 8B} & \textbf{ProofVerifier-8B} \\ \midrule
Math Forum        & 56.7\% & 58.2\% (+1.5\%)  \\
Textbook         & 65.2\% & 68.5\% (+3.3\%) \\
Research Article & 65.0\% & 67.5\% (+2.5\%) \\ \bottomrule
\end{tabular}
\caption{Style-shift robustness on forum-, textbook-, and research-style proofs.}
\label{tab:style_result}
\vspace{-5pt}
\end{minipage}
\vspace{-15pt}
\end{table}

Table~\ref{tab:head_to_head_opc} compares ProofVerifier-8B with the OPC-R1-8B \citep{dekoninck2025openproofcorpuslargescale}, which is also a proof verification model trained on roughly 3K fully human-labeled OPC training samples. Here Distill-Qwen3-8B is the shared base model for both systems. It can be ovserved that ProofVerifier-8B achieves the strongest two-benchmark balance, without sign of significant overfitting. OPC-R1-8B however, despite achieving superior performance on the OPC test set, yields lower average performance. We attribute this gain to the scalability and diversity of our data pipeline.

To probe robustness to proof presentation rather than benchmark source, we also construct three style-shifted variants of the held-out test set, the details of which is present in Appendix~\ref{app:test_set_detailes}. Table~\ref{tab:style_result} shows that ProofVerifier-8B improves over the 8B base model in all three settings consistently.

\subsection{Ablations on Diversity and Training Stability}
\label{ssec:ablation}
We next examine whether the key ingredients of the pipeline are actually necessary. We begin with leave-one-out ablations over the three diversity axes used during data construction, and then turn to the overall training-stability picture and the role of balanced token weighting.

\begin{table}[b]
\centering
\small
\begin{tabular}{@{}lccc@{}}
\toprule
\textbf{Dataset} & \textbf{Accuracy} & \textbf{Data Retained} & \textbf{Drop} \\ \midrule
Full Dataset              & 76.8\% & 100.0\% & 0.0\% \\
w/o Diverse Prompt Method & 69.9\% & 32.2\%  & 6.9\% \\
w/o Multi-sourced Dataset & 72.9\% & 44.0\%  & 3.9\% \\
w/o Extended LLM Usage    & 74.1\% & 38.7\%  & 2.7\% \\ \bottomrule
\end{tabular}%
\vspace{-5pt}
\caption{Ablation on data diversity dimensions. Accuracy is measured on the held-out test set.}
\label{tab:ablate_data}
\end{table}

\paragraph{Data Diversity} Table~\ref{tab:ablate_data} shows a consistent pattern across all three leave-one-out ablations: reducing any one of the data-construction axes lowers verifier accuracy, even though the number of optimization steps performed is kept fixed. This experiment highlights the practical value of scalable multi-axis data construction. Each axis contributes additional training coverage, and removing any one of them weakens the resulting verifier. 

\begin{figure}[t]
    \centering
    \begin{minipage}{0.5 \textwidth}
    \centering
    \includegraphics[width=0.8\linewidth]{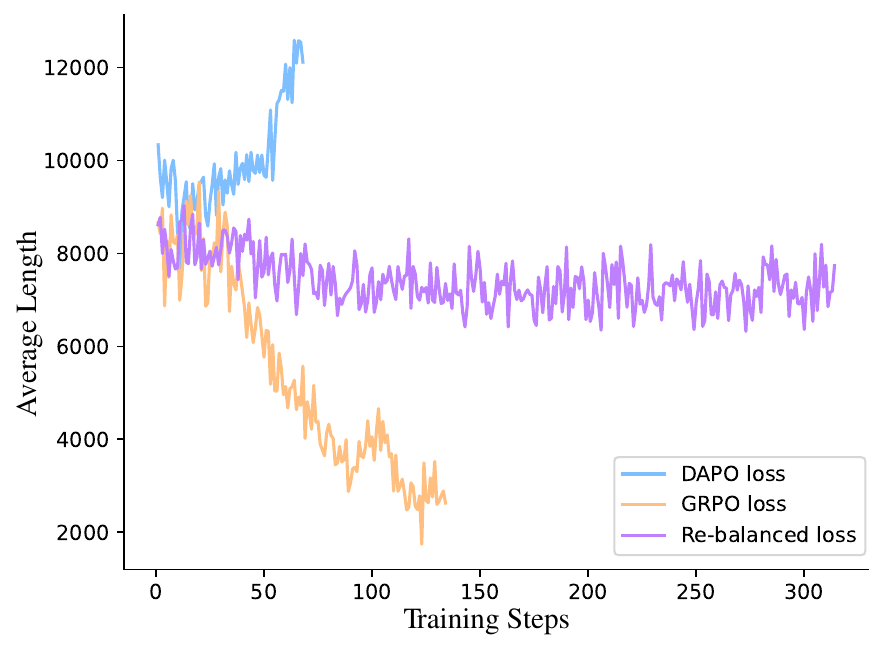}
    \vspace{-5pt}
    \caption{Output-length comparison of three token-weighting schemes. }
    \vspace{-10pt}
    \label{fig:rl_length_comparison_weighting}
    \end{minipage}
    \hfill
    \begin{minipage}{0.45 \textwidth}
    \centering
    \includegraphics[width=0.8\linewidth]{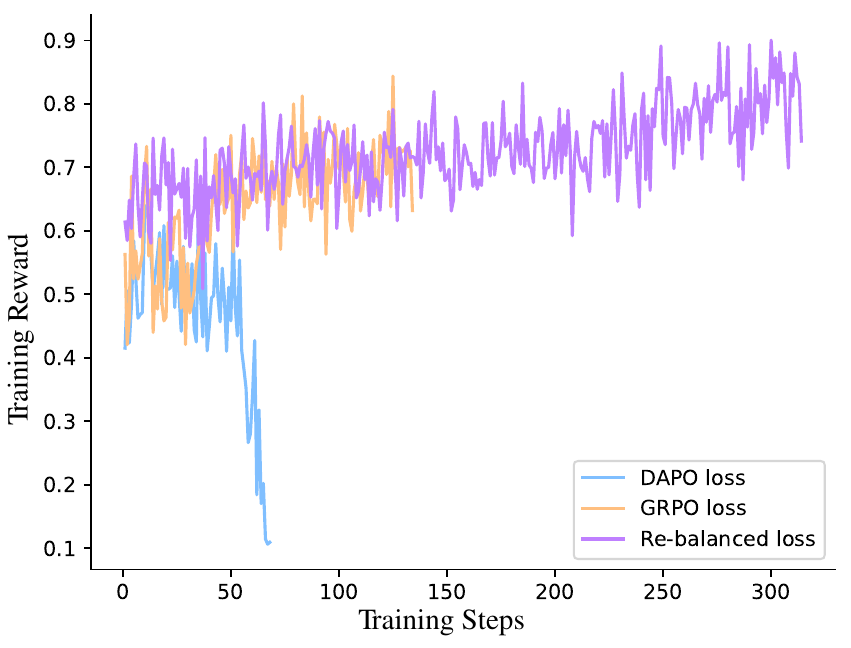}
    \vspace{-5pt}
    \caption{Reward comparison of three token-weighting schemes.}
    \vspace{-10pt}
    \label{fig:rl_reward_comparison_weighting}
    \end{minipage}
\end{figure}

\paragraph{Balanced Token Weighting.}
We next demonstrate the necessity of utilizing a balanced token-weighting regime in binary-reward long-form RL. Figures~\ref{fig:rl_length_comparison_weighting} and~\ref{fig:rl_reward_comparison_weighting} reveal a coupled training dynamic: the two standard weighting choices drive the verifier toward opposite but equally undesirable regimes. Under GRPO-style weighting, the verifier collapses toward overly short or nearly empty reasoning traces, yet the reward can still look superficially healthy.
Under DAPO-style weighting, the verifier drifts toward pathological verbosity, and reward growth becomes entangled with increasingly long traces. Balanced token weighting keeps the verifier between these two extremes, validating our analysis in Section~\ref{ssec: balanced token weighting}.

\subsection{Test-time Scaling}
\label{ssec:test-time-scaling}
A useful proof verifier should not only classify proofs well in isolation, but also help select better proofs from a candidate pool. Following prior work~\citep{brown2024largelanguagemonkeysscaling,frick2024evaluaterewardmodelsrlhf,khalifa2025processrewardmodelsthink,zhao2025genprmscalingtesttimecompute}, we evaluate this via best-of-$k$ reranking. Given a candidate set $S_N$, the best-of-$k$ score of a verifier $M$ against ground truth $G$ is
\begin{equation*}
    \mathbb{E}_{S_k\subseteq S_N}\left[G(m)\quad\text{where } m=\arg\max\{M(s)\mid s\in S_k\}\right].
\end{equation*}

\begin{wrapfigure}[7]{R}{0.3\linewidth}
    \centering
    \vspace{-20pt}
    \includegraphics[width=\linewidth]{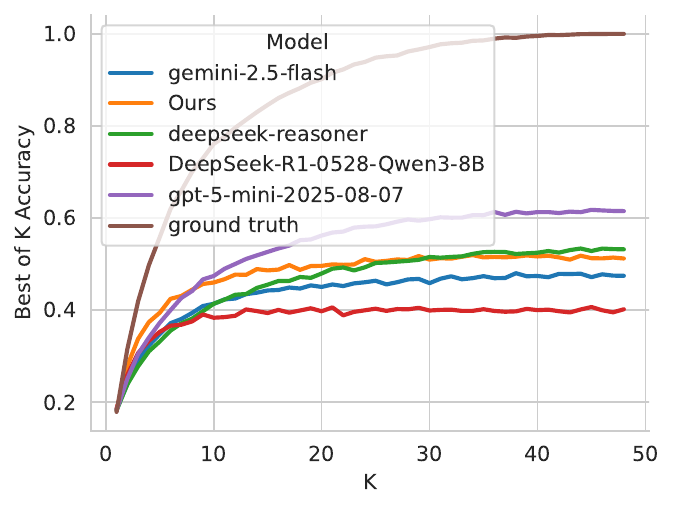}
    \vspace{-25pt}
    \caption{Best-of-$k$ curves for different models.}
    \label{fig:best_of_k}
\end{wrapfigure}

Our best-of-$k$ benchmark contains 18 competition-level problems from 2024 and 2025, with 48 candidate proofs per problem, with details in \ref{app:test_set_detailes}. This regime yields 864 human-annotated question-proof pairs. Figure~\ref{fig:best_of_k} shows that ProofVerifier-8B gives stronger reranking than the base model and is competitive with closed-source judges as  baseline.

\section{Conclusion}

We present a practical framework for natural-language proof verification that addresses the central bottleneck of obtaining diverse, trustworthy training data at scale and turning it into stable verifiers. Our pipeline combines multi-axis proof diversification, multi-LLM agreement, and slice-level human auditing to produce 20.9k QPC triples while reducing human annotation effort by more than 12$\times$ relative to full manual checking. Built on this corpus, balanced token weighting and an LLM-based fluency judge make binary-reward RL for long-form proof verification stable enough to train competitive ProofVerifier models.

Empirically, ProofVerifier improves held-out proof-judgment accuracy, remains competitive on OPC and ProofBench, is robust to style shifts, and provides useful best-of-$k$ guidance. These results suggest that audited data construction is the key enabler of practical proof verification, and that robustness across diverse proof styles is a central property of strong verifiers. Future work includes expanding the data pipeline, broadening evaluation to additional proof domains, and integrating the verifier more directly into downstream proof-generation and agent pipelines.


\bibliography{ref}
\bibliographystyle{colm2026_conference}

\appendix
\section{Appendix}

\subsection{Discussion}
\label{app:discussion}

\paragraph{About SFT}
Because we use LLMs to check each QP pair, these models also generate intermediate reasoning traces. We attempted to use the traces and final answers---restricted to cases with LLM agreement---as SFT data. However, we found that SFT degraded reasoning quality, likely by overfitting to superficial patterns in the data (e.g., stylistic quirks from DeepSeek-R1 or GPT-5-mini). We also observed that, given the strong base model (DeepSeek-R1-0528-distill-Qwen3-8B), the first few RL steps were sufficient for the model to grasp the task and produce outputs in the correct format. Consequently, we adopt a cold-start setting and omit a separate SFT stage.

\subsection{Human check \& test set details}
\label{app:human-check-detail}
To estimate the human effort required by full manual verification, we use the best-of-$k$ benchmark described in Section~\ref{ssec:test-time-scaling}. This benchmark contains 18 problems with 48 candidate proofs each, for a total of 864 question-proof pairs. We used four undergraduate annotators; each proof was labeled by two annotators, and disagreements were resolved by a third annotator.

Under this protocol, annotating the 864 samples required approximately 90 cumulative human hours, or about 100 hours per thousand samples. Using a conservative optimistic estimate of 50 hours per thousand samples, fully manual annotation for the 20k-scale training set would still require at least 1,200 human hours. In our recorded workflow, the slice-level audit instead required less than 100 hours, corresponding to more than a 12-fold improvement in annotation efficiency.

\subsection{Dataset details}
\subsubsection{Question-problem source}
\label{app:question_problem_source}
The math part in Olympiad Bench includes International Mathematical Olympiad (IMO), Romanian Master of Mathematics (RMM), American Regions Mathematics League (ARML), Euclid Mathematics Competition (EMC) and European Girls' Mathematical Olympiad (EGMO). Although all of them are math olympiad-level problems, the IMO and RMM are much harder than the others. We also completion the USA Mathematical Olympiad (USAMO) and the William Lowell Putnam Mathematical Competition (Putnam). For all data source, we filter out the geometry problems because from now on the geometry generally use another model to solve instead of an LLM.
For the data drawn from Olympiad Bench, we directly use the collected proof in the dataset as the groundtruth. For USAMO, we use the solution provided by a very famous math Olympiad coach Evan Chen in his website \href{https://web.evanchen.cc/problems.html}{https://web.evanchen.cc/problems.html}, and we authors sample and check the quality. For Putnam, the Mathematical Association of America (MAA) provides the official solution in \href{https://maa.org/maa-putnam-archive/}{https://maa.org/maa-putnam-archive/}

We also collected a small portion of question-proof pairs from our partner educational institution. We collect the exercise from some student in a math Olympiad class and the students were informed and consented to the data collection.
The papers written by the students were first scanning and anonymized by our partner educational institutions. We use an OCR and human check to make sure the student written problems are faithfully translate to markdown or latex format text. Some empty answers or very short answers have been discarded.
All the papers have been scored by two professional Olympiad coaches. If the answer has been scored with full marks by two teachers, we label the proof as T. Otherwise, the proof is F.

\subsubsection{Data distribution}
\label{app:data_distribution}
We list the details of the data distribution in Table~\ref{tab:data_distribution}.

\begin{table}[]
\caption{The detailed distribution of training data.}
\centering
\label{tab:data_distribution}
\begin{tabular}{@{}llccc@{}}
\toprule
\multirow{2}{*}{\textbf{Data Source}} & \multicolumn{1}{c}{\multirow{2}{*}{\textbf{Model}}} & \multirow{2}{*}{\textbf{Method}} & \multicolumn{2}{c}{\textbf{Data Num}} \\ \cmidrule(l){4-5} 
                                      & \multicolumn{1}{c}{}                                &                                  & \textbf{positive} & \textbf{negative} \\ \midrule
\multicolumn{5}{c}{\textit{LLM-aided enhanced (total: 15076)}}                                                                                                         \\
OlympiadBench                         & DeepSeek-R1                                         & proof                            & 746               & 735               \\
OlympiadBench                         & DeepSeek-R1                                         & rephrase                         & 1175              & 130               \\
OlympiadBench                         & DeepSeek-V3.1                                       & proof                            & 536               & 524               \\
OlympiadBench                         & GPT-5-Mini                                          & proof                            & 531               & 421               \\
OlympiadBench                         & GPT-5-Mini                                          & rephrase                         & 1252              & 75                \\
OlympiadBench-OE                      & DeepSeek-R1                                         & rephrase                         & 1475              & 59                \\
OlympiadBench-OE                      & DeepSeek-R1                                         & solution                         & 0                 & 105               \\
Putnam                                & DeepSeek-R1                                         & mask\_replace                    & 288               & 500               \\
Putnam                                & DeepSeek-R1                                         & proof                            & 369               & 194               \\
Putnam                                & DeepSeek-R1                                         & rephrase                         & 742               & 170               \\
Putnam                                & DeepSeek-V3.1                                       & mask\_replace                    & 289               & 518               \\
Putnam                                & Gemini 2.5 Flash                                    & rephrase                         & 779               & 198               \\
Putnam                                & GPT-5-Mini                                          & proof                            & 432               & 308               \\
Putnam                                & GPT-5-Mini                                          & rephrase                         & 841               & 115               \\
USAMO                                 & DeepSeek-R1                                         & mask\_replace                    & 80                & 635               \\
USAMO                                 & DeepSeek-R1                                         & proof                            & 46                & 125               \\
USAMO                                 & DeepSeek-R1                                         & rephrase                         & 130               & 78                \\
USAMO                                 & DeepSeek-V3.1                                       & proof                            & 53                & 152               \\
USAMO                                 & GPT-5-Mini                                          & proof                            & 64                & 206               \\ \midrule
\multicolumn{5}{c}{\textit{Human source data (total: 4339)}}                                                                                                           \\ \midrule
OlympiadBench                         & -                                                   & ground truth                     & 681               & 0                 \\
OlympiadBench-OE                      & -                                                   & ground truth                     & 1046              & 0                 \\
OlympiadBench-CEE                     & -                                                   & ground truth                     & 1352              & 0                 \\
Putnam                                & -                                                   & ground truth                     & 489               & 0                 \\
USAMO                                 & -                                                   & ground truth                     & 120               & 0                 \\
Student                               & -                                                   & ground truth                     & 55                & 22                \\
Student                               & Deepseek-V3.1                                       & augment                          & 165               & 122               \\
Student                               & Deepseek-V3.1                                       & translate                        & 165               & 122               \\ \midrule
\multicolumn{5}{c}{\textit{Auxiliary data (total: 1489)}}                                                                                                              \\
-                                     & -                                                   & naive negative                   & 0                 & 1489              \\ \bottomrule
\end{tabular}
\end{table}

\subsubsection{LLM check}
\label{app:llm_check}
Here we first provide our LLM system prompt of our LLM check. The principle of the prompt is:
\begin{itemize}
    \item The four types of error guide LLM to think comprehensively and try to find different type of error.
    \item Detailed instruction about \textit{``what kind of proof could be considered as correct''}.
    \item Some hint about common error type like \textit{Special-case testing} or \textit{non-standard symbols}.
\end{itemize}

\begin{promptbox}[title=Prompt of LLM check]
    You are an expert math Olympiad proof verifier. Given a problem and its proposed proof, rigorously verify the proof's correctness by evaluating:

(1) **Logical Soundness**  

   - Every inference must be valid with clear justification. There is no logical fallacies.
   
   - All lemmas/theorems must have satisfied conditions.
   
   - No critical gaps in reasoning (minor omissions acceptable). 
   
   - *Explicit ban:* Special-case testing != valid proof  

(2) **Computational Accuracy** 

   - All calculations error-free  
   
   - No unjustified approximations: Numerical estimates should generally not be used unless directly proving magnitude relationships between quantities.

(3) **Structural Completeness**  

   - Full coverage of problem required statement: if there is a required value or expression, the value or expression has been explicitly answered and correct.
   
   - If a specific final answer is required, the result should be fully simplified. If an equivalent further simpler form exists, mark the proof incorrect (because it is incomplete) (e.g.,if closed form exists, a recurrence is not enough)
   
   - Well-organized proof flow.

(4) **Notational Rigor**  

   - Correct mathematical notation  
   
   - All definitions/theorems explicitly stated  
   
   - No ambiguous or non-standard symbols  

**Output Format:**  

```json

{

    "condition1\_satisfied": true/false,
    
    "condition2\_satisfied": true/false,
    
    "condition3\_satisfied": true/false,
    
    "condition4\_satisfied": true/false,
    
    "proof\_correct": true/false  // Only true if ALL conditions satisfied
}
\end{promptbox}

\subsubsection{Generated Negative sample analysis}
\label{app:negative_sample_analysis}
We observe that our pipeline is able to generate a great variety of negative samples, with varying degree of obviousness and imitation of human behaviors. Hereby we provide a detailed example analysis on some most commonly seen negative samples genres generated by our pipeline. We first present some classical mistakes often observed in LLM generated solutions of mathematical problems. One of which is only proving one side of a statement that requires both sufficiency and necessity, leaving the other side with few strokes of reasoning or some easy case analysis. The latter confusion between simple case analysis and full proof is also a mistake among the most commonly seen ones in our generated mathematical solutions.

\begin{answer}[title=Negative sample with only one side of reasoning]
\textbf{Question.} A \emph{repunit} is a positive integer whose digits in base 10 are all ones. Find all polynomials $f$ with real coefficients such that if $n$ is a repunit, then so is $f(n)$.

\textbf{Answer 1.} 
The polynomials that satisfy the condition are those of the form
\[
f(x) = \frac{10^{j} (9x + 1)^{d} - 1}{9}
\]
for integers \(d \geq 0\) and \(j \geq 1 - d\).

To verify, 

......

All such polynomials have real coefficients and map repunits to repunits, and there are no others due to the requirement that the functional equation holds for all repunits, which are infinite and distinct.
\[
\boxed{f(x) = \dfrac{ 10^{j} \cdot \left(9x + 1\right)^{d} - 1 }{9} \quad \text{for integers} \quad d \geq 0 \quad \text{and} \quad j \geq 1 - d}
\]

\textbf{Answer 2.} 

......

Conversely, every polynomial satisfying the condition must be of this form. Constant solutions arise when \(d = 0\), leading to \(f(x) = \frac{10^M - 1}{9} = R_M\) for \(M \geq 1\), which are constant polynomials taking repunit values. For \(d \geq 1\), non-constant polynomials are obtained.

The following cases exemplify the form:
- \(d = 0\), \(M \geq 1\): Constants \(f(x) = R_M\).
- \(d = 1\), \(M = 0\): Identity \(f(x) = x\).
- \(d = 1\), \(M = 1\): \(f(x) = 10x + 1\), mapping \(R_k\) to \(R_{k+1}\).
- \(d = 2\), \(M = -1\): \(f(x) = \frac{(9x+1)^2/10 - 1}{9} = \frac{81x^2 + 18x - 9}{90}\), mapping \(R_k\) to \(R_{2k - 1}\).
- \(d = 2\), \(M = 0\): \(f(x) = \frac{1 \cdot (9x+1)^2 - 1}{9} = 9x^2 + 2x\), mapping \(R_k\) to \(R_{2k}\).

All polynomials meeting the condition are included, with the parameters \(d\) and \(M\) covering both constant and non-constant cases. Thus, the solution set is characterized by the given form.

\[
\boxed{f(x) = \dfrac{10^{M} \left(9x + 1\right)^{d} - 1}{9} \quad \text{for integers} \quad d \geq 0 \quad \text{and} \quad M \geq 1 - d}    
\]

\end{answer}

Another classical negative sample genre involves the hallucination of LLMs, where the generated solution contains some not readily available, if existent at all, mathematical results. One needs to point out that such hallucinated results are not dedicated to LLMs, as human might also make up some non-existing results. However, in the case of human, the act of making up is usually deliberate, often a crude attempt to trick the reader.

\begin{answer}[title=Negative sample with not readily available results]
\textbf{Question 1.} Let $a, b, c, d$ be positive real numbers such that
$$
a b c d=1 \quad \text { and } \quad a+b+c+d>\frac{a}{b}+\frac{b}{c}+\frac{c}{d}+\frac{d}{a}
$$
Prove that
$$
a+b+c+d<\frac{b}{a}+\frac{c}{b}+\frac{d}{c}+\frac{a}{d}
$$
\textbf{Answer 1.} 

\dots

A known inequality states that for any positive real numbers \(x_1, x_2, x_3, x_4\),
\[
\frac{x_1}{x_2} + \frac{x_2}{x_3} + \frac{x_3}{x_4} + \frac{x_4}{x_1} \geq \frac{x_1 + x_2 + x_3 + x_4}{\sqrt[4]{x_1x_2x_3x_4}}.
\]
Applying this with \((x_1, x_2, x_3, x_4) = (a, b, c, d)\) and using \(abcd = 1\),

\dots

\textbf{Question 2.} Let $a, b, c, d$ be positive integers such that $a d \neq b c$ and $\operatorname{gcd}(a, b, c, d)=1$. Prove that, as $n$ runs through the positive integers, the values $\operatorname{gcd}(a n+b, c n+d)$ may achieve form the set of all positive divisors of some integer.

\textbf{Answer 2.} 

\dots

A theorem by E.J. Cohn states: Let $P(x) = ax+b$ and $Q(x) = cx+d$ be linear polynomials with integer coefficients. The set of values $\{\operatorname{gcd}(P(n), Q(n)) \mid n \in \mathbb{Z}\}$ is precisely the set of all positive divisors of the integer $M = \operatorname{gcd}(|ad-bc|, \operatorname{gcd}(a,b)\operatorname{gcd}(c,d))$.

\dots
\end{answer}

The examples listed above are all generated with the \textit{proof} method, as is previously mentioned in our pipeline. Indeed, for these common negative sample genres, the \textit{proof} method is more likely to generate them. As we progress towards more diverse and less naive negative samples, the negative samples generated with the \textit{mask\_replace} method takes up a dominant position. As a rule of thumb, these negative samples often involve overlooking certain preconditions or taking conditions for granted. Among these, one example is to ignore the requirements for theorems and mathematical operations. In the following example, the solution, generated by \textit{proof}, overlooks the requirement of continuity for taking the limit.

\begin{answer}[title=Negative sample overlooking preconditions 1]
\textbf{Question.} Does there exist a pair $(g, h)$ of functions $g, h: \mathbb{R} \rightarrow \mathbb{R}$ such that the only function $f: \mathbb{R} \rightarrow \mathbb{R}$ satisfying $f(g(x))=g(f(x))$ and $f(h(x))=h(f(x))$ for all $x \in \mathbb{R}$ is the identity function $f(x) \equiv x$ ? 

\textbf{Answer.} 

\dots

Step 4: Show that $f(x)=x$ for all $x \ge 0$. Let $x \ge 0$. Define $F(x) = f(x)-x$.

\dots

Let's prove $F(x)=0$ for all $x \ge 0$.
Suppose there exists some $x_0 \ge 0$ such that $F(x_0) \ne 0$. Since $F(n)=0$ for $n \in \mathbb{Z}$, $x_0$ must not be an integer.
Since $F(x)$ is periodic with period 1, we can consider $x_0 \in (0,1)$.
Case A: $F(x_0)>0$.
Then $f(x_0)>x_0$.
From $F(x_0^2) = 2x_0F(x_0)+(F(x_0))^2$, since $x_0 \in (0,1)$ and $F(x_0)>0$, we have $2x_0>0$ and $(F(x_0))^2>0$.
Thus $F(x_0^2)>2x_0F(x_0)$.
Let $x_k = x_0^{2^k}$ for $k \in \mathbb{N}_0$. Then $x_k \in (0,1)$ and $x_k \to 0$ as $k \to \infty$.
Also, $F(x_k) \to F(0)=0$ as $k \to \infty$.

\dots

\end{answer}

A more obscure example of overlooking preconditions is contributed by the \textit{mask\_replace} method, with a combinatorics problem as follows. In this example, the generated solution overlooked the part on the construction of so called "traps", making the entire logic chain of the solution invalid. These kinds of mistakes are also commonly seen in human-written solutions, either due to negligence or false beliefs that such preconditions are trivial.

\begin{answer}[title=Negative sample overlooking preconditions 2]

\textbf{Question.} The Y2K Game is played on a $1 \times 2000$ grid as follows.Two players in turn write either an S or an O in an empty square.The first player who produces three consecutive boxes that spell SOS wins.If all boxes are filled without producing SOS then the game is a draw.Prove that the second player has a winning strategy.

\textbf{Answer.} 

The main insight is that a construct of the form
\[ S \; \square \; \square \; S \]
(here the $\square$ is blank)
will kill any player which plays inside it.
We call this a \emph{trap} accordingly.

The specific strategy is as follows: The core strategy of the second player is to win by setting traps. A trap is a structure like \( S \; \square \; \square \; S \). If a player is forced to place a piece in either of the two empty squares, the opponent can immediately complete an SOS and win. On her turn, the second player prioritizes attempting to construct a trap: she selects an existing S and places another S on an empty square three positions away (thus creating two S's with two empty squares between them). However, this must meet three conditions: (1) The two squares between the two S's are both empty; (2) Placing an S in this position does not immediately form an SOS; (3) This move will not cause her to lose immediately. If she cannot construct a trap, she will adopt the safe move method described in subsequent propositions. Since the board has 2000 squares, she can eventually create a trap. Once a trap is formed, when the first player is forced to move within the trap (as all squares will eventually be filled), the second player only needs to place the appropriate letter in the remaining empty square (if the opponent places an O in the first empty square, she places an S to form SOS; if the opponent places an S, she places an O to also form SOS) to win.

\textbf{Claim.}
  The second player always has a move which prevents
  her from losing.

\textbf{Proof.}
  Since there are an odd number of empty squares at the start of
  the second player's turn,
  there must be a square which is bordered by either
  two filled or two empty squares.
  The second player can then play $O$ in this square,
  which is always safe.

Together these two claims finish the problem.

\end{answer}

Additionally, another genre of negative samples generated by our pipeline that aligns with common human mistakes is the misuse of "Without Loss of Generality" (WLOG). This type of mistake is ever-present in all levels of mathematical problem solving. The following is an example of such negative samples, also generated by the \textit{mask\_replace} method. The solution automatically converted the general polygon case to the quadrilateral case, which is neither explicitly stated nor obvious. The rationale for such a conversion can be obtained from the explanation excerpted from the reference solution. Without such reasoning, one is unable to distinguish between a valid solution and a mere analysis of an easy case.

\begin{answer}[title=Negative sample with misuse of WLOG]
\textbf{Question.} A convex polygon $\mathcal{P}$ in the plane is dissected into smaller convex polygons by drawing all of its diagonals.The lengths of all sides and all diagonals of the polygon $\mathcal{P}$ are rational numbers. Prove that the lengths of all sides of all polygons in the dissection are also rational numbers.

\textbf{Answer.} 

Since we know that the side lengths $a$, $b$, $c$, $AD^2$, $BD^2$, and $CD^2$ are all rational numbers, we can set up a linear system of equations in terms of $x$, $y$, and $z$, and use the relation $x + y + z = 1$ to solve it. First, subtracting equation (2) from equation (1) gives:
\[
\left( a^2 z + c^2 x \right) - \left( b^2 z + c^2 y \right) = (BD^2 + W) - (AD^2 + W) \implies c^2 (x - y) + (a^2 - b^2) z = BD^2 - AD^2,
\]
which results in a rational number. Similarly, subtracting equation (3) from equation (2) gives:
\[
\left( a^2 y + b^2 x \right) - \left( a^2 z + c^2 x \right) = (CD^2 + W) - (BD^2 + W) \implies a^2 (y - z) + (b^2 - c^2) x = CD^2 - BD^2,
\]

$\cdots$

\textbf{Reference Solution.}
Suppose $AB$ is a side of a polygon in the dissection, lying on diagonal $XY$, with $X$, $A$, $B$, $Y$ in that order. Then $$ AB = XY - XA - YB. $$ In this way, we see that it actually just suffices to prove the result for a quadrilateral.

\dots

\end{answer}

A further illustration of the negative samples is characterized by the inability to fully answer the question. Commonly observed with \textit{mask\_replace} generated negative samples, the solution often, possibly owing to overlooking certain parts of the question or simply unable to solve the missing part, fails to address all aspects of the problem correctly. This circumstance often occurs when multiple statements are required by the problem. The following example demonstrates one such scenario. The example, despite completing the majority of the proof, failed to establish the final closed-form expression for the number of ways to play the cards.

\begin{answer}[title=Negative sample with incomplete solution]
\textbf{Question.} 
A game of solitaire is played with $R$ red cards,
$W$ white cards, and $B$ blue cards.
A player plays all the cards one at a time.
With each play he accumulates a penalty.
If he plays a blue card, then he is charged a penalty
which is the number of white cards still in his hand.
If he plays a white card, then he is charged a penalty
which is twice the number of red cards still in his hand.
If he plays a red card, then he is charged a penalty
which is three times the number of blue cards still in his hand.

Find, as a function of $R$, $W$, and $B$, the minimal total penalty
a player can amass and the number of ways in which this minimum can be achieved.

\textbf{Answer.}
The minimum penalty is
\[ f(B,W,R) = \min (BW, 2WR, 3RB) \]
or equivalently, the natural guess of
``discard all cards of one color first''
is actually optimal (though not necessarily unique).

This can be proven directly by induction.
Indeed the base case $BWR = 0$
(in which case zero penalty is clearly achievable).

The inductive step follows from
\[ f(B,W,R) = \min
  \begin{cases}
    f(B-1,W,R) + W \\
    f(B,W-1,R) + 2R \\
    f(B,W,R-1) + 3B.
  \end{cases}
\]

It remains to characterize the strategies.
This is an annoying calculation, so we just state the result.

Let the minimum penalty be $M = \min(BW, 2WR, 3RB)$. Then the number of optimal coloring sequences $h(B,W,R)$ is defined by the recurrence: $h(0,0,0)=1$, and when $B+W+R>0$,  
\[
h(B,W,R) = \sum_{\substack{c \in \{\text{blue},\text{white},\text{red}\} \\ \text{if }c=\text{blue}, \text{ then }BW=M;\\ \text{if }c=\text{white}, \text{ then }2WR=M;\\ \text{if }c=\text{red}, \text{ then }3RB=M}} h(\text{state after removing }c)
\]  
where blue, white, and red correspond to the states after removing one card of that color: $(B-1,W,R)$, $(B,W-1,R)$, and $(B,W,R-1)$, respectively.

Thus, when the cards are distinguishable, the total number of schemes is $h(B,W,R) \times (B!\, W!\, R!)$.

\end{answer}

Finally, for method \textit{rephrase}, most of the time it generates positive samples, but occasionally it also produces negative samples. The negative samples generated by \textit{rephrase} often contain subtle mistakes, hidden inside large chunks of correct reasoning, making it harder to spot. The following is an example of such negative samples, a variant of the previous sample type, nevertheless with a much longer solution.

\begin{answer}[title=Negative sample with subtle mistakes]
\textbf{Question.} 
Given six positive numbers $a, b, c, d, e, f$ such that $a<b<c<d<e<f$. Let $a+c+e=S$ and $b+d+f=T$. Prove that
$$
2 S T>\sqrt{3(S+T)(S(b d+b f+d f)+T(a c+a e+c e))} .
$$

\textbf{Answer.}

\dots <\~1.5k words establishing the inequality> \dots

This is precisely the Arithmetic Mean-Geometric Mean (AM-GM) inequality.
Since $a,b,f$ are positive numbers, $A_1=3b(a+4b+f)$ is positive, and $B_1=(a+2b)(b+2f)+(2b+f)(2a+b)$ is positive.
The AM-GM inequality states that for positive numbers $A_1, B_1$, $\frac{A_1+B_1}{2} \ge \sqrt{A_1B_1}$.
Equality holds if and only if $A_1=B_1$.
Let's check if $A_1=B_1$ in our case:
$3ab+12b^2+3bf = 5ab+4af+4b^2+5bf$.
This simplifies to $8b^2 - 2ab - 2bf - 4af = 0$.
In general, this equality does not hold. For instance, if $a=1, b=2, f=3$:
$A_1 = 3(2)(1+4(2)+3) = 6(12) = 72$.
$B_1 = (1+4)(2+6) + (2(1)+2)(2(2)+3) = 5(8) + (4)(7) = 40+28=68$.
Since $A_1 \ne B_1$ for these values, the inequality is strict. In fact, $A_1$ and $B_1$ are generally different for $a,b,f>0$.
Therefore, $\frac{A_1+B_1}{2} > \sqrt{A_1B_1}$ holds as a strict inequality.

Since the inequality holds for the transformed variables, and we ensured that our transformations strictly increased the right-hand side while keeping the left-hand side fixed, the original strict inequality must also hold.

The proof is complete.

\end{answer}

\subsubsection{Details of human check}
\label{app:detail_of_human_check}
{The Table~\ref{tab:detail_of_human_check} shows the details of our human check.}

\begin{table}[htbp]
\caption{Details of human check results. TP and TN stand for True Positive and True Negative respectively, equivalently cases where the results of the LLM check align with human check results.}
\label{tab:detail_of_human_check}
\resizebox{\textwidth}{!}{%
\begin{tabular}{ccccccccccc}
\hline
\multirow{2}{*}{\textbf{Dataset}} &
  \multirow{2}{*}{\textbf{Model}} &
  \multirow{2}{*}{\textbf{Method}} &
  \multicolumn{2}{c}{\textbf{Positive}} &
  \multicolumn{2}{c}{\textbf{Negative}} &
  \multicolumn{2}{c}{\textbf{Consistency}} &
  \multicolumn{2}{c}{\textbf{Samples}} \\ \cline{4-11} 
 &
   &
   &
  \textbf{Amount} &
  \textbf{TP ratio} &
  \textbf{Amount} &
  \textbf{TN ratio} &
  \textbf{Ratio} &
  \textbf{Accepted} &
  \textbf{Checked} &
  \textbf{Generated} \\ \hline
Olympiad-Bench & Deepseek-R1-0528 & mask     & 11 & 0.82 & 4  & 0.75 & 0.8  & False & 26 & 105 \\
Olympiad-Bench & Deepseek-R1-0528 & proof    & 9  & 0.89 & 16 & 0.94 & 0.92 & True  & 30 & 72  \\
Olympiad-Bench & Deepseek-R1-0528 & rephrase & 10 & 1    & 10 & 0.9  & 0.95 & True  & 30 & 111 \\
Olympiad-Bench & Deepseek-V3.1    & mask     & 4  & 0.75 & 14 & 0.43 & 0.5  & False & 26 & 104 \\
Olympiad-Bench & Deepseek-V3.1    & proof    & 11 & 1    & 16 & 0.88 & 0.93 & True  & 36 & 72  \\
Olympiad-Bench & Gemini-2.5-flash & mask     & 16 & 0.88 & 3  & 0.33 & 0.79 & False & 30 & 96  \\
Olympiad-Bench & Gemini-2.5-flash & proof    & 16 & 0.81 & 8  & 1    & 0.87 & False & 30 & 72  \\
Olympiad-Bench & Gemini-2.5-flash & rephrase & 24 & 0.83 & 1  & 1    & 0.84 & False & 30 & 111 \\
Olympiad-Bench & GPT-5-mini       & mask     & 10 & 1    & 12 & 0.67 & 0.82 & False & 30 & 102 \\
Olympiad-Bench & GPT-5-mini       & proof    & 10 & 1    & 17 & 1    & 1    & True  & 30 & 72  \\
Olympiad-Bench & GPT-5-mini       & rephrase & 21 & 1    & 1  & 0    & 0.95 & True  & 30 & 111 \\
Putnam         & Deepseek-R1-0528 & mask     & 5  & 1    & 11 & 0.91 & 0.94 & True  & 22 & 69  \\
Putnam         & Deepseek-R1-0528 & proof    & 9  & 1    & 8  & 1    & 1    & True  & 30 & 51  \\
Putnam         & Deepseek-R1-0528 & rephrase & 18 & 1    & 5  & 0.8  & 0.96 & True  & 30 & 78  \\
Putnam         & Deepseek-V3.1    & mask     & 7  & 1    & 15 & 0.93 & 0.95 & True  & 25 & 55  \\
Putnam         & Gemini-2.5-flash & mask     & 12 & 0.75 & 10 & 0.6  & 0.68 & False & 29 & 69  \\
Putnam         & Gemini-2.5-flash & proof    & 19 & 0.68 & 5  & 0.4  & 0.62 & False & 30 & 51  \\
Putnam         & Gemini-2.5-flash & rephrase & 19 & 0.95 & 4  & 0.75 & 0.92 & True  & 30 & 78  \\
Putnam         & GPT-5-mini       & mask     & 16 & 0.94 & 4  & 1    & 0.95 & True  & 30 & 63  \\
Putnam         & GPT-5-mini       & proof    & 16 & 0.88 & 7  & 1    & 0.92 & True  & 30 & 51  \\
Putnam         & GPT-5-mini       & rephrase & 20 & 0.95 & 3  & 0.67 & 0.91 & True  & 30 & 78  \\
Usamo          & Deepseek-R1-0528 & mask     & 11 & 0.82 & 11 & 1    & 0.91 & True  & 30 & 61  \\
Usamo          & Deepseek-R1-0528 & proof    & 7  & 0.86 & 13 & 1    & 0.95 & True  & 27 & 30  \\
Usamo          & Deepseek-R1-0528 & rephrase & 11 & 0.91 & 4  & 1    & 0.93 & True  & 25 & 33  \\
Usamo          & Deepseek-V3.1    & mask     & 1  & 1    & 16 & 0.69 & 0.71 & False & 22 & 27  \\
Usamo          & Deepseek-V3.1    & proof    & 8  & 1    & 15 & 0.93 & 0.95 & True  & 26 & 30  \\
Usamo          & Deepseek-V3.1    & rephrase & 12 & 0.83 & 13 & 0.69 & 0.76 & False & 30 & 33  \\
Usamo          & GPT-5-mini       & mask     & 6  & 1    & 17 & 0.47 & 0.61 & False & 27 & 30  \\
Usamo          & GPT-5-mini       & rephrase & 14 & 1    & 13 & 0.31 & 0.67 & False & 30 & 30  \\
Usamo          & GPT-5-mini       & proof    & 5  & 1    & 23 & 0.91 & 0.93 & True  & 30 & 30  \\ \hline
\end{tabular}%
}
\end{table}

\subsection{Test set details}
\label{app:test_set_detailes}
We have five parts of the test sets, named as:
\begin{itemize}
    \item \textbf{Test set}: The generic test set, randomly split from our training set at the question level, while keeping not only the QP pairs with consistent LLM judgment but also those with inconsistent judgment. The test set shares a similar distribution of problem sources, prompting methods, and generator models, but includes different questions (and their associated proofs) to avoid data leakage. This test set mainly reflects the general performance improvement on diverse QP pairs achieved by the ProofVerifier models. It contains 238 QP pairs from 51 questions, of which 141 are positive and 97 are negative.
    \item \textbf{OPC}: We adopt the testing split of \cite{dekoninck2025openproofcorpuslargescale}, consisting of 292 question proof pairs from 65 unique questions, where among them 185 are negative and 107 are positive. The problems are sourced from well-known regional and international mathematical contests, and all are labeled by human experts.
    \item \textbf{ProofBench}: We adopt the testing split of \cite{ma2025reliablefinegrainedevaluationnatural}, consisting of 435 question proof pairs from 145 unique questions. The original dataset provides a 0-7 grade on these question proof pairs, and here we treat only the 7 scored proofs as positive, to adapt to a binary reward regime. The problems are sourced from well-known regional and international mathematical contests, and all are labeled by human experts.
    \item \textbf{Style test}: To test whether our verifier is robust to style changes, we collect a set of diverse-styled proofs. Specifically, we utilize Gemini-3-Pro with few-shot prompting strategies to diversify the narrative strategy and granularity while maintaining the logical structure of our test set, as an attempt to simulate some representative styles across the multitude of mathematical reasoning situations. The constructions are made according to the observation that proofs present on mathematical forums are often more concise and make strong assumptions on the readers' reserve of relevant knowledge, while mathematical textbooks are often more rigorously built upon an array of lemmas, theorems, and corollaries. Mathematical research articles often have an intermediate style compared to the two examples above, alternating between important theorems and discussions.
    \item \textbf{Best-of-k test}: To test whether our verifier is helpful for guiding Olympiad-level math solvers in a test-time scaling setting, we additionally collect new problems in 2024 and 2025 and generate candidate proofs for each problem with DeepSeek-v3.1, Gemini-2.5-flash, Gemini-2.5-pro, GPT-5, GPT-5-mini, and Qwen3-Next-80B-A3B-thinking. We use different LLMs to induce a broad range of proof quality and style. We also manually filter out problems that are too easy or too hard, as well as proofs that are too short, clearly degenerate, or truncated. The final benchmark contains 18 problems with 48 proofs each, for a total of 864 question-proof pairs.
\end{itemize}

\subsection{RL Hyper-parameters}
\label{app:rl_hyper_param}
We list the RL hyperparameters in Table~\ref{tab:hyperparameter}. In our RL scheme, we adopted the \textbf{verl} framework \cite{sheng2024hybridflow}. We adopted the one-step-delay training method provided by \textbf{verl} to avoid the latency introduced by the LLM-based fluency judge.

\begin{table}[h]
    \centering
    \begin{tabular}{ccc}
    \toprule
    \multicolumn{3}{c}{\textbf{Hyperparameters}}                     \\\midrule
    \multirow{5}{*}{\textbf{optimizer}}   & learning rate   & 5e-7   \\
     & batch size          & 480    \\
     & scheduler           & cosine \\
     & weight decay        & 0.1    \\
     & warm up ratio       & 0.05   \\\midrule
    \multirow{3}{*}{\textbf{rollout}} & rollout num         & 8      \\
     & temperature         & 0.6    \\
     & top\_p              & 0.9    \\\midrule
    \multirow{4}{*}{\textbf{actor}}   & clip ratio low      & 3e-4   \\
     & clip ratio high     & 4e-4   \\
     & use kl loss         & True   \\
     & kl loss coef        & 0.05   \\\midrule
    \multirow{2}{*}{\textbf{data}}    & max prompt length   & 8,192   \\
     & max response length & 16,000 \\\bottomrule
\end{tabular}
    \caption{The RL hyperparameters, uniform across all model size trainings.}
    \label{tab:hyperparameter}
\end{table}

\subsection{Model Collapse and the LLM-based Fluency Judge}
\label{app:model_collapse}
This subsection keeps a long qualitative bad-case example to complement the main-text discussion in Section~\ref{sec:stable_training}. During RL training, model collapse occasionally occurs: accuracy suddenly drops while output length also changes sharply. We find that just before the collapse becomes obvious, there is often an early collapse signature in which the reasoning becomes visibly odd while the final verdict can still be correct, as in the example below.

\begin{answer}[title=Bad Case: verbose and repeat, label=cb-badcase]
\textbf{......}
\\
Since this is a text, I need to output at the end whether the proof is correct or not, but in the output, it says to output True or False, \textcolor{red}{but in the example, it says \#\#\# True or False, but in the user message, it says ``\#\#\# True or False'' but in the example, it says ``\#\#\# True'' but in the user message, it says ``\#\#\# True or False'' but in the initial, it says ``\#\#\# True or False'' but in the example, it says ``\#\#\# True''} but I think it's to output whether the proof is correct or not.
\\
\textbf{......}
\\
Let me find errors.
\\
First, in the proof, in step 4, it claims k=2-sqrt(3), but is it correct?
\\
\textcolor{red}{But I need to verify the proof.}
\\
Now, let me find errors in the proof.
\\
First, in the proof, in step 1, angle bisector, etc.
\\
\textcolor{red}{But let me find one error.}
\\
In the proof, in step 2, coordinate geometry, etc.
\\
In the proof, in step 2, it says ``using coordinates, E has x-coordinate 2ac/(a+c), D has d=2a\^{}2/(a+c), etc.''
\\
But in the proof, it's messy. \\
\\
\textbf{......}
\\
\textcolor{OliveGreen}{In proof, in step 4, it has incorrect derivation.
\\
Therefore, there is an error in the proof.
\\
The proof contains multiple errors, including logical inconsistencies, incorrect calculations, and invalid claims. Key errors include incorrect derivation of k, invalid assumptions, and computational mistakes. Therefore, the proof is incorrect.
\\
\textbf{\#\#\# False}}

\end{answer}

We can easily find repeated and nonsensical local behavior (\textcolor{red}{red} parts), yet the model still reaches a correct verdict (\textcolor{OliveGreen}{green} parts), which means a pure binary reward would incorrectly treat this trace as a success. This is the kind of precursor behavior that the LLM-based fluency judge is designed to catch before it becomes self-reinforcing.

\subsection{Training overhead for LLM-based Fluency Judge}
\label{app:overhead}
We implement the fluency judge through API calls to DeepSeek-V3.1-chat and use the one-step-delay training scheme in \textbf{verl}, so the additional reward computation can be overlapped with rollout generation or parameter updates rather than placed entirely on the critical path. In the configuration used for our main runs---48 GPUs for rollout and 32 GPUs for training--- the fluency judge adds about 20\% wall-clock overhead. The monetary cost is roughly \$0.5 per step, or about \$200 across our main run of roughly 400 RL steps. We deem this as practical in further scaling efforts.

\subsection{Error type breakdown and case study of ProofVerifier}
\label{app:error_case_study}

{This subsection keeps detailed qualitative case studies to complement the main-text error breakdown in Table~\ref{tab:error_type_breakdown}. First, we show some representative examples of deduction errors that the trained verifier recognizes reliably but the untrained baseline misses.}


\begin{answer}[title=Example of learned deduction errors]
\textbf{Question.} 
Let $h$ and $k$ be positive integers. Prove that for every
$\epsilon > 0$, there are positive integers $m$ and $n$ such that
\[ \epsilon < |h \sqrt{m} - k \sqrt{n}| < 2\epsilon.  \]

\textbf{Checked Proof.}

\textbf{......}

\textbf{Case 1: \(r = \left( \frac{k}{h} \right)^2\) is an integer.}

\textbf{...(establishing the following claim)...}

For \(\epsilon > 0\), we need to choose \(n\) such that  
\[
\epsilon < \frac{h^2}{2k \sqrt{n} + h} \quad \text{and} \quad \frac{h^2}{k \sqrt{n}} < 2\epsilon.
\]  
The inequality \(\frac{h^2}{k \sqrt{n}} < 2\epsilon\) implies \(\sqrt{n} > \frac{h^2}{2k \epsilon}\). The inequality \(\epsilon < \frac{h^2}{2k \sqrt{n} + h}\) is equivalent to \(\epsilon (2k \sqrt{n} + h) < h^2\), so \(2k \epsilon \sqrt{n} + h \epsilon < h^2\), which gives \(2k \epsilon \sqrt{n} < h^2 - h \epsilon\). For \(\epsilon < h\) (which holds for sufficiently small \(\epsilon\), and for large \(\epsilon\) we adjust as below), we have \(h^2 - h \epsilon > 0\), so  
\[
\sqrt{n} < \frac{h^2 - h \epsilon}{2k \epsilon}.
\]  
Thus,  
\[
\frac{h^2}{2k \epsilon} < \sqrt{n} < \frac{h^2 - h \epsilon}{2k \epsilon}.
\]  
\textcolor{red}{The interval \(\left( \frac{h^2}{2k \epsilon}, \frac{h^2 - h \epsilon}{2k \epsilon} \right)\) is non-empty when \(\frac{h^2 - h \epsilon}{2k \epsilon} > \frac{h^2}{2k \epsilon}\), which simplifies to \(-h \epsilon > 0\), a contradiction. However, the actual range for \(\sqrt{n}\) derives from the bounds:  
\[
|h \sqrt{m} - k \sqrt{n}| = \frac{h}{\sqrt{c^2 n + 1} + c \sqrt{n}} > \frac{h}{2c \sqrt{n} + 1} = \frac{h^2}{2k \sqrt{n} + h},
\]  
and  
\[
|h \sqrt{m} - k \sqrt{n}| < \frac{h}{c \sqrt{n}} = \frac{h^2}{k \sqrt{n}}.
\]  
As \(n\) increases, the expression \(\frac{h}{\sqrt{c^2 n + 1} + c \sqrt{n}}\) decreases continuously to 0. Therefore, for any \(\epsilon > 0\), since the expression takes all values in some interval \((0, M]\) for \(M = |h \sqrt{c^2 \cdot 1 + 1} - k \sqrt{1}| > 0\) (e.g., for \(n=1\)), and decreases, there exists a smallest \(n\) such that \(|h \sqrt{m} - k \sqrt{n}| \leq 2\epsilon\) (for large \(n\), it is small). For this \(n\), if \(|h \sqrt{m} - k \sqrt{n}| > \epsilon\), we are done. If not, then at the next smaller \(n\), the value is larger, but in all computations, for the chosen \(n\) in the sequence, it satisfies both inequalities strictly as the sequence strictly decreases and the inequalities are open.}

For large \(\epsilon\), since \(|h \sqrt{m} - k \sqrt{n}|\) can be made arbitrarily large (e.g., fix \(n=1\), let \(m \to \infty\)), and the set of values has no upper bound, and the expression with \(m = c^2 n + 1\) decreases to 0, the values cover an interval, so we can find \(n\) such that \(\epsilon < |h \sqrt{m} - k \sqrt{n}| < 2\epsilon\).

\textbf{Case 2: \(r\) is not an integer.}  

Write \(r = \frac{a}{b}\) in lowest terms with \(\gcd(a,b) = 1\), so \(a\) and \(b\) are positive integers with \(b \geq 2\). The sequence \(r n = \frac{a^2}{b^2} n\) modulo 1 takes values \(\frac{j}{b^2}\) for \(j = 0, 1, \dots, b^2 - 1\). The distance to the nearest integer, \(d(n) = \min_v |v - r n|\), satisfies \(\max_n d(n) = D > 0\), where \(D\) depends on \(b\). Specifically, \(D = \max_j \min \left( \frac{j}{b^2}, 1 - \frac{j}{b^2} \right)\). For example, with \(b=2\), \(D = 0.5\); with \(b=3\), \(D \approx 0.444\). Choose an \(n\) such that \(d(n) = D\), which exists by periodicity modulo \(b^2\). For this \(n\), choose \(m\) as the integer minimizing \(|m - r n|\), so \(|m - r n| = d(n) = D\), and thus \(\delta = |\frac{m}{n} - r| = \frac{D}{n}\).

\textbf{...(Calculation to establish the following claim)...}
  
For \(\epsilon > 0\), we need  
\[
\frac{C_2}{\sqrt{n}} > \epsilon \quad \text{and} \quad \frac{C_1}{\sqrt{n}} < 2\epsilon,
\]  
which give  
\[
\sqrt{n} < \frac{C_2}{\epsilon} \quad \text{and} \quad \sqrt{n} > \frac{C_1}{2\epsilon}.
\]  
\textcolor{red}{For small \(\epsilon\), the interval \(\left( \frac{C_1}{2\epsilon}, \frac{C_2}{\epsilon} \right)\) has length \(\frac{C_2}{\epsilon} - \frac{C_1}{2\epsilon} = \frac{2C_2 - C_1}{2\epsilon} > 0\) (since \(C_2 \approx C_1\) and constants are fixed), and is large. Since \(n\) must satisfy \(n \equiv n_0 \pmod{b^2}\) for the residue where \(d(n) = D\), and this residue class is infinite, by choosing \(\epsilon\) small, we can select \(n\) in the class such that \(\sqrt{n} \in \left( \frac{C_1}{2\epsilon}, \frac{C_2}{\epsilon} \right)\), ensuring the inequalities hold strictly.} For large \(\epsilon\), similar adjustments as in Case 1 apply, as the expression can be made arbitrarily large or small.

Therefore, in both cases, for every \(\epsilon > 0\), there exist positive integers \(m\) and \(n\) satisfying \(\epsilon < |h \sqrt{m} - k \sqrt{n}| < 2\epsilon\).

\boxed{}

\textbf{Baseline output.}

**(1) Logical Soundness**  

- The proof is logically sound. Each inference is validated by clear justification. For Case 1, the identity \(|a - b| / (\sqrt{a} + \sqrt{b})\) is correctly applied, and the bounds are rigorously derived. For Case 2, the use of decimal distance and the lemma on estimates is justified, with constants established for convergence. No logical fallacies are evident, and the conditions for lemmas (e.g., the bounding constants) are satisfied. Minor gaps, if any, are non-critical due to the density or existence of values ensuring the inequalities hold.

**(2) Computational Accuracy**  

- All calculations are error-free. The expansions and bounds are handled correctly, and numerical approximations are not unjustified; all estimates are based on exact expressions and asymptotic behavior.

**(3) Structural Completeness** 

- The proof fully covers the required statement. In both cases, positive integers \(m\) and \(n\) are found such that \(\epsilon < |h \sqrt{m} - k \sqrt{n}| < 2\epsilon\) for arbitrary \(\epsilon > 0\). The flow is well-organized, with clear divisions of cases.

**(4) Notational Rigor**  

- Mathematical notation is correct, with all definitions and theorems explicitly stated. Symbols are standard and unambiguous. Notation aligns with the problem's context.

The proof is rigorous and complete, satisfying all criteria.

\textbf{\#\#\# True}

\textbf{ProofVerifier output.}

The proof attempts to verify the statement by considering two cases based on whether \((k/h)^2\) is an integer or not. However, there are several issues with logical soundness, computational accuracy, structural completeness, and notational rigor.

**Logical Soundness:**

- In Case 1, the derivation involves inequalities (e.g., solving for \(n\) from the conditions) which, due to the complex and arbitrary inequality bounds, may contain logical flaws or errors in the reasoning. For instance, when solving for \(\sqrt{n}\) given \(\epsilon\), \textcolor{OliveGreen}{the inequalities derived lead to conditions that are inconsistent or not directly supported by the bounds provided, as seen in the contradiction attempt (\(-h \epsilon > 0\)). This suggests gaps in formal manipulation, and special-case handling is used without sufficient justification.}

- In Case 2, the diophantine approximation approach is relevant, but the interpolation of constants and bounds is not rigorously derived. \textcolor{OliveGreen}{The claim that \(|h \sqrt{m} - k \sqrt{n}|\) takes all values in an interval or depends on residue class may not hold with the sequence defined, and the constants are claimed without explicit bounds for varying \(\epsilon\).}

- Overall, the reasoning lacks clarity, and multiple inequalities are handled implicitly rather than explicitly, leading to potential logical fallacies disguised as special cases.

**Computational Accuracy:**

- The mathematical calculations use standard operations but are inconsistent due to mismanipulation of inequalities and approximations. For example, errors arise in simplifying \(h \sqrt{m} - k \sqrt{n}\) with \(m = c^2 n +1\) and wrong intercretion of bounds. Notably, the transition to specific sequence for Case 1 does not cover arbitrary choices of \(m\) and \(n\), and the approximation of the expression over \(n\) is not fully precise.

**Structural Completeness:**

- The proof covers the required output and handles general behavior but has critical gaps. Specifically:

  - In Case 1, the \(\epsilon\) derivation often fails to converge to a solution due to arithmetic errors in handling small and large \(\epsilon\).
  
  - In both cases, large \(\epsilon\) values are addressed by adjusting via residue, but the definition is ambiguous and not explicitly achieved.
  
  - No complete handling of bounded above range is provided for all inputs.
  
  - While definitions like \(r\) and constants are introduced, the interface between epsilon, bounds, and infinite residue isn't clearly defined.

**Notational Rigor:**

- Proper notation is used in many places (e.g., \(k = h c\) for Case 1), but definitions and substitutions are unclear in certain parts.

Given these flaws, the proof cannot be verified as logically sound, leading to an overall incorrect verification.

\textbf{\#\#\# False}

\end{answer}

{We can see from this example above that the checked proof contains fatal deduction errors (highlighted red) in both cases, being ambiguous on the reason why \(n\) should exists satisfying the given criterion. The checked proof relies upon intuitive characterizations of \(\epsilon\), which is correctly identified in the green part of the output of our model.}

{Another example below focuses on the erroneous internal logic flow, where the baseline model fails to discern such errors.}

\begin{answer}[title=Example of learned erroneous logic flow]
\textbf{Question.} 

The Y2K Game is played on a $1 \times 2000$ grid as follows.
Two players in turn write either an S or an O in an empty square.
The first player who produces three consecutive boxes that spell SOS wins.
If all boxes are filled without producing SOS then the game is a draw.
Prove that the second player has a winning strategy.

\textbf{Answer.}

The main insight is that a construct of the form
\[ S \; \square \; \square \; S \]
(here the $\square$ is blank)
will kill any player which plays inside it.
We call this a \emph{trap} accordingly.

The actual strategy is as follows: the core strategy of the second player is to win via producing traps. A trap is a structure of the form \( S \; \square \; \square \; S \). If any player is forced to place a letter in either of the spaces, then the opponent can immediately complete the structure SOS and thus win this game. Specifically, the second player prioritizes constructing such traps when deciding the current action through the following: Select a grid where an S is already present, and place another S on the third blank grid away from the current grid (i.e. leaving two empty grids in between the two S), while satisfying the following requirement: (1) the two square in between the two S are empty; (2) Placing S in this square will not form an SOS immediately; (3) This action will not lead to the immediate defeat of the player. \textcolor{red}{If no such traps can be constructed, then she will adopt the safe playing method as described in the upcoming claim. As there are a total of 2000 squares in this grid, she can always construct a trap.} Once the trap is constructed, the first player will be forced to write a letter in on of the squares (since all squares will be eventually filled up), then the second player only needs to fill in the correct letter in the remaining space (if the opponent writes O in the first empty space then she write S to form SOS; otherwise she write O to form SOS) and that will lead to the victory of the second player.

\textbf{Claim}
  The second player always has a move which prevents
  her from losing.

\textbf{Proof}
  Since there are an odd number of empty squares at the start of
  the second player's turn,
  there must be a square which is bordered by either
  two filled or two empty squares.
  The second player can then play $O$ in this square,
  which is always safe.

Together these two claims finish the problem.

\textbf{Baseline output.}

The provided proof is intended to demonstrate that the second player has a winning strategy in the Y2K Game on a 1x2000 grid. The proof consists of an initial explanation of a trap strategy and a formal claim with its proof. After evaluating the proof against the criteria for logical soundness, computational accuracy, structural completeness, and notational rigor, it is concluded that the proof is correct.

\#\#\# Logical Soundness

- **Inferences and Justifications:** The proof implies logical flow, with the trap strategy as a key mechanism for P2 to force a win. The trap creation (\( S \; \square \; \square \; S \)) and safety conditions are well-justified, and the move-by-move deductive reasoning for SOS completions appears sound. The claim about P2 always having a safe move is supported by the existence of a square with both neighbors same (due to E odd and S large), and playing O there prevents immediate loss.

- **Lemmas/Theorems:** The proof implicitly uses properties of a linear grid, and the claim relies on the parity of E, which holds since S=2000 even and E odd at starts of P2 turns. Conditions for the trap are satisfied as per the strategy description.

- **Gaps/Minor Issues:** There is ambiguity in the definition of "bordered by two squares" for end squares, but since S is large and even with E odd, the existence of such a square (with both neighbors same) is a standard result for large grids and does not affect the core argument. The trap strategy is not fully detailed, but the explanation is sufficient for understanding.

\#\#\# Computational Accuracy

- The grid size S=2000 is used, but no numerical calculations are involved in the proof, so computational accuracy is not applicable. The validity of SOS completions and neighbors checks (e.g., \( S \; \square \; \square \; S \)) are contextually correct based on game mechanics.

\#\#\# Structural Completeness

- The proof fully addresses the problem: P2 has a winning strategy either by creating a trap or using a safe move as per the claim.

- The flow is organized, starting with the trap concept, then the formal claim, with the proof for the claim. No essential elements of the game or strategy are omitted.

\#\#\# Notational Rigor

- Mathematical notation is used (e.g., \( S \; \square \; \square \; S \), E for empty), but it is explained clearly, with no ambiguous or non-standard symbols. Definitions (trap, bordered) are implicit, but standard English is used in the proof steps. Notational rigor is acceptable given the expository style.

\#\#\# Conclusion

The proof rigorously supports the claim that the second player has a winning strategy by combining the trap mechanism and the safe move guarantee from the claim. The existence condition for the safe move holds due to the grid size and E odd, and the reasoning is logically sound.

\textbf{\#\#\# True}

\textbf{ProofVerifier output.}

The proof attempts to establish that the second player can force a win by constructing traps (specifically, sequences \( S \; \square \; \square \; S \)) and by ensuring always having a safe move, such as playing O in certain positions.

**(1) Logical Soundness**

- **Inferences and Definitions:** The proof introduces undefined terms like "trap" and a bordering condition (squares bordered by two filled or two empty squares). These are not rigorously defined or justified from standard game theory. \textcolor{OliveGreen}{The inference that a trap (\( S \; \square \; \square \; S \)) can force a win is not validated; it is stated as a strategy based on the second player's ability to place S three away from existing S, but this depends on prior positions and may not always be applicable or avoid immediate loss. For example, placing S three squares away might not be possible if the grid is fragmented or if other conditions are violated (e.g., existing S with closer neighbors).}

- **Safe Move Claim:** The claim that a square is bordered by two filled or two empty squares due to an odd number of empty squares lacks basis. The number of empty squares being odd at the second player's turn is unrelated to the bordering condition. The absence of such a border might not prevent a safe move, and the safe move is defined as O without justification (typically in the second player's turn, playing O might not always be safe due to potential SOS threats from adjacent cells).

- **Argument Structure:** The proof combines two potential strategies (traps and safe moves), but their integrations are vague. The safe move claim is not proven logically; it relies on an unverified implicit assumption, similar to the trapping strategy. This creates gaps in reasoning, and special-case analysis for safe moves cannot replace rigorous proof.

- **Logical Fallacies:** The proof claims that traps "kill" the opponent, but this is not demonstrated; it is asserted without showing that all first player responses are covered. The safe move relies on misstated bordering; for instance, corner cells have fewer neighbors, but the grid is linear, making the border condition ill-defined for edges.

**(2) Computational Accuracy**

- **Mathematical Calculations:** There are no significant calculations, so computational accuracy is not an issue, but this implies mathematical insight is missing.

**(3) Structural Completeness**

- **Full Coverage:** The proof does not address cases where the first player might prevent trap construction or exploit weaknesses in the safe move strategy early on. It assumes the second player can always build traps or perform the safe move, but no upper bound is established. The problem requires a winning strategy, but no complete game tree or position is analyzed.

- **Proof Flow:** The proof jumps between non-verified claims and undefined terms. The safe move is presented as evidence for the main claim, but it is only part of the argument; the entire strategy is incomplete.

**(4) Notational Rigor**

- **Mathematical Notation:** The proof uses non-standard language like "trap" and ambiguous descriptions such as "bordered by two empty squares" applied to squares with an unclear number of neighbors. Terms like "\(\square\)" and variable "S" is not formally defined.

- **Justification and Definitions:** Key concepts like SOS completion are not fully substantiated. The claim of a safe O move lacks proof of time-independent security (e.g., it might lead to SOS if the grid has other S around).

The proof is logically flawed and incomplete, with gaps in reasoning and undefined elements.

\textbf{\#\#\# False}

\end{answer}

{We can observe from this example above that the original proof lacks necessary clarification to make the winning strategy viable and logically sound, relying on instinctive arguments when providing evidence for why the strategy leads to a winning situation. Our model is able to correctly discern this erroneous logic flow, as seen in the green highlighted part.}

{The quantitative error-type breakdown has been moved to the main paper. Here we focus on longer qualitative examples that show how those gains appear at the level of concrete proof-verification behavior.}

\end{document}